\documentclass[final,3p,times]{elsarticle}
\usepackage{graphicx}
\usepackage{amssymb}
\usepackage{amsmath}
\usepackage[utf8]{inputenc}
\usepackage{multirow}
\usepackage{times}
\usepackage{latexsym}
\usepackage{url}
\usepackage{subcaption}
\usepackage{lineno,hyperref}
\usepackage{todonotes}
\usepackage{geometry}
\usepackage{pdflscape}

\newtheorem{definition}{Definition}[section]

\journal{Computer Speech and Language}
\begin{document}

\begin{frontmatter}

\title{Comprehensive Analysis of Aspect Term Extraction Methods using \\ Various Text Embeddings}
% \title{LSTM-based Character Embedding model and other Word Representation Techniques for Aspect Term Extraction - Comprehensive Analysis}

\tnotetext[mytitlenote]{Fully documented templates are available in the elsarticle package on \href{http://www.ctan.org/tex-archive/macros/latex/contrib/elsarticle}{CTAN}.}

\author[wrut]{Łukasz Augustyniak\corref{cor}}
\ead{lukasz.augustyniak@pwr.edu.pl}

\author[wrut]{Tomasz Kajdanowicz}
\ead{tomasz.kajdanowicz@pwr.edu.pl}

\author[wrut]{Przemysław Kazienko}
\ead{kazienko@pwr.edu.pl}

%% or include affiliations in~footnotes:
\address[wrut]{Department of Computational Intelligence, Wrocław University of Science and Technology, Wrocław, Poland}

\cortext[cor]{Corresponding author}

\begin{abstract}
Recently, a~variety of model designs and methods have blossomed in the context of the sentiment analysis domain. However, there is still a~lack of comprehensive studies of Aspect-based Sentiment Analysis. We want to fill this gap and propose a~comparison with ablation analysis of Aspect Term Extraction using various text embeddings methods. We particularly focused on simple architectures based on long short-term memory (LSTM) with optional conditional random field (CRF) enhancement using different pre-trained word embeddings. Moreover, we analyzed the influence on the performance of extending the word vectorization step with character-based word embeddings. The experimental results on SemEval datasets revealed that bi-directional long short-term memory (BiLSTM) could be used as a very good predictor, even comparing to very sophisticated and complex models using huge word embeddings or language models. We presented a comprehensive analysis of various customizations of LSTM-based architecture and word/character embeddings that could be used as a guideline to choose the best model version for particular user needs. 

\end{abstract}  


\end{frontmatter}

% \linenumbers

\section{Introduction}

If you have used Uber, TripAdvisor or Amazon, you are among 100 million (Uber), 450 million (TripAdvisor), or over 300 million (Amazon) active users. All of these businesses provide services with a~strong focus on communication and a~relationship with customers. It is fundamental for their success to listen to their clients, understand what exactly the customer is saying and engage when it is necessary. However, how can we analyze even a~glimpse of these communications? This is a~reason why development of natural language processing methods (NLP) for large amounts of such data has boomed. Analysis of textual data can provide valuable insights by the processing of direct feedback from the customers (customer reviews or their complaints) found on social media platforms such as Twitter, Facebook and many more platforms, where people regularly post their opinions on all kinds of businesses. Hence, what kind of NLP techniques should we apply to extract useful knowledge from opinionated texts? The standard sentiment analysis methods annotate the whole texts or documents with one class only such as negative, positive or neutral.  However, it would be helpful to narrow and precisely describe the insights described in texts. There exists a sub area of sentiment analysis called  Aspect-based Sentiment Analysis (ABSA). 

\begin{definition}[Aspect-based Sentiment Analysis]
Aspect-based sentiment analysis aims to extract the sentiment polarity of the document toward the specific \textit{aspect} (also called \textit{attribute}) of a~given more general concept.
\end{definition}

\begin{figure}[!ht]
\centering
\includegraphics[scale=0.1]{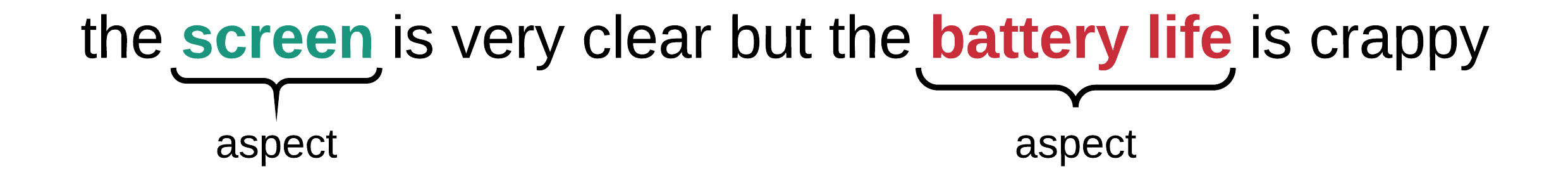}
\caption{Example of aspects in a phone review.}
\label{fig:aspect-example}
\end{figure}

Imagine that we have a~phone review such as in Figure \ref{fig:aspect-example}. We can spot positive sentiment polarity for the \textit{screen} aspect and negative polarity for the \textit{battery life}. Unfortunately, nowadays most of the solutions still use sentiment analysis only on the whole document level; hence, they can not distinguish between the sentiment polarity related to \textit{screen} and \textit{battery life}. They commonly treat the document as a~source of only one opinion. Uber would be interested in which aspect of their service is rated positively and which negatively. There is a~big difference between opinion about the mobile app and a~driver - they are described by two different aspect sets. 

\begin{figure}[!ht]
\centering
\includegraphics[scale=0.11]{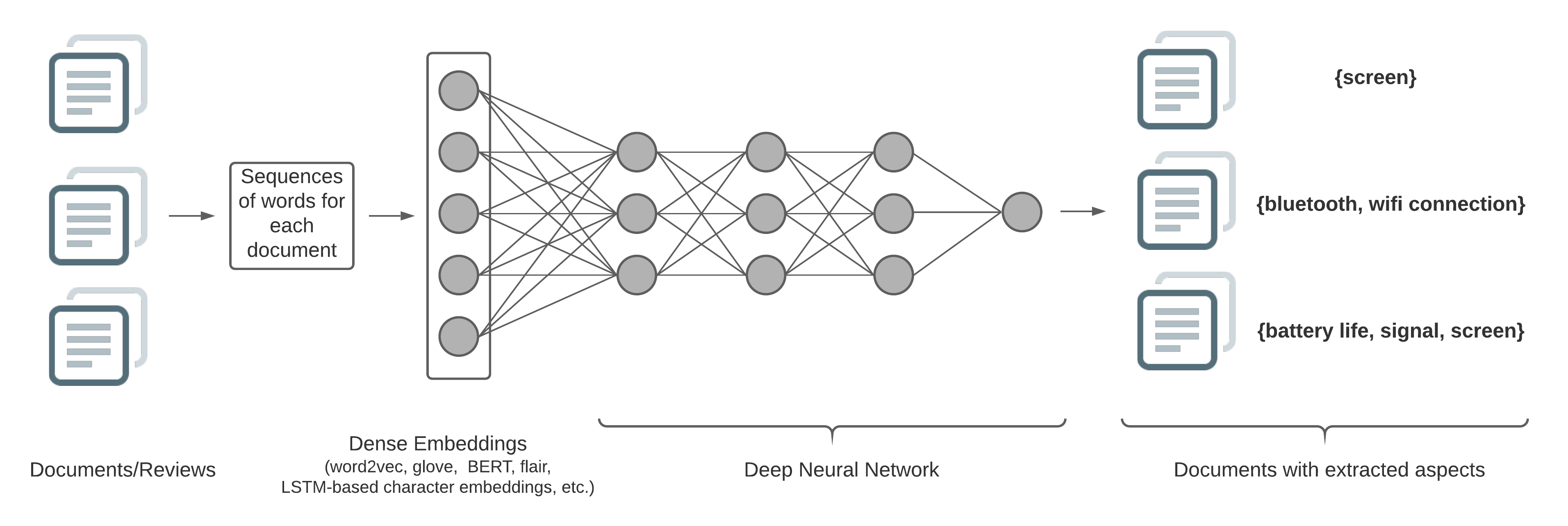}
\caption{A bird's eye view of Aspect Term Extraction from a phone opinions.}
\label{fig:aspect-term-extraction-general-flow}
\end{figure}

An aspect-based sentiment analyzer consists of many components. The first and primary one is responsible for precise and complete Aspect Term Extraction (ATE). Why is this step is so crucial? Aspect term extraction has a~substantial influence on the accuracy of the entire sentiment analysis tool because errors at the beginning (input) of the whole pipeline will be propagated to the next steps and could potentially harm the entire solution seriously~\cite{10.5555/3121646.3121651}. The Aspect Term Extractor takes some documents as input and identifies a set of aspects for each document. Figure \ref{fig:aspect-term-extraction-general-flow} presents an overview of such extractors with neural network architecture.

There has been increasing emphasis on neural network architectures, and sequence tagging approaches \cite{Poria2016, double-embeddings, Wehrmann2017-twitter, Barnes2017} in Aspect Term Extraction. These techniques have proved to be effective for named entity recognition (NER),  part-of-speech tagging (POS) or chunking tasks \cite{Lample2016, Ma2016, Akbik2018}. There exists some research presenting neural network-based models for aspect term extraction, but there is still a lack of comprehensive and reliable analysis of them. They mostly cover only a~limited range of solutions, use only one or two pre-trained word embeddings~\cite{Li2017, Poria2016, double-embeddings}; focus only on very well represented languages such as English~\cite{Ruder2016, Zhou}; compare only a few model configurations, e.g., two word embedding models and two neural network architectures~\cite{Yin:2016:UWD:3060832.3061038, Li2017}, or fine-tune models with hand-crafted, language-dependent rules \cite{Poria2016}. Nowadays, machine learning models are deployed in many solutions, with different constraints. Researchers and machine learning engineers must consider aspects such as model complexity, memory requirements, the need for GPU acceleration, and many more. A single performance metric is not always enough. 

Our goal is to address the mentioned above problems proposing an end-to-end Aspect Term Extraction model in comparison with a comprehensive ablation analysis. We want to present a guideline that will make the decision of choosing a particular Aspect Term Extraction architecture easier for researchers or industry machine learning engineers. Our experiments cover more than ten diverse word embeddings extracted from different corpora and using different models. We propose the extension of pre-trained word embeddings with character embeddings to improve models of underrepresented languages such as Polish. 

\newpage
In our analysis, we responded to the following research questions:

\subsection*{Research Question 1 - Robustness}
\label{sec:rs1}
\textbf{How robust are the general\footnote{Pre-trained based on general texts such Wikipedia, Common Crawl, etc.} word embeddings in domain-dependent Aspect Term Extraction?}

\subsection*{Research Question 2 - Coverage}
\label{sec:rs2}
\textbf{How does the coverage of word embeddings impact the performance of Aspect Term Extraction?}

\subsection*{Research Question 3 - Character-based Word Embeddings}
\label{sec:rs3}
\textbf{When are character-based word embedding methods able to eradicate drawbacks of the static pre-trained word embeddings in Aspect Term Extraction?}

\subsection*{}

Summing up, our main contributions are (1) a~new method for Aspect Term Extraction using both word and character embeddings (LSTM-based embeddings), (2) a~comprehensive comparison of a number of LSTM-based approaches to ATE based on many pre-trained word embeddings, and (3) an ablation analysis with focus on what is the influence of the text vectorization methods and model characteristics on the final performance.

\section{Related Work}
\label{sec:related_work}

In this section, we present the most popular machine learning approaches to Aspect Term Extraction as well as methods for text vectorization.

\subsection{Machine Learning approaches to Aspect Term Extraction}

One of the first ideas for the solution of aspect extraction in a supervised learning manner was the use of a linear chain Conditional Random Field (CRF). For example, Toh and Wang proposed such an approach in their DLIREC system~\cite{Toh2014} or Chernyshevich~\cite{Chernyshevich2014} from IHS R\&D. The other often used model was SVM, as in \cite{alvarezlopez-EtAl:2016:SemEval, Brun2014}. There exist also mixed models combining supervised learning and rule-based systems as in \cite{Poria2014, Nguyen2015, Poria2016}. Jakob and Gurevych \cite{Jakob:2010:EOT:1870658.1870759} proposed using a sequence tagging scheme for aspect extraction. They used features such as token information, POS, short dependency path, word distance, and information about opinionated sentences. Toh and Wang \cite{Toh2014} extended this approach with more hand-crafted features such as lexicons, syntactic and semantic features, as well as cluster features induced from unlabeled data.

Nowadays, deep learning-based approaches have emerged recently. Recursive neural networks and conditional random field has been used by Want et al.~\cite{D16-1059}. Poria et al.~\cite{Poria2016} proposed a~deep convolutional neural network that tags each word in the document as either an aspect or non-aspect word (sequence tagging approach). Nevertheless, they also used hand-crafted linguistic patterns to improve their method. Hai et al.~\cite{10.1007/978-3-319-57529-2_28} used a convolution stacked neural network using dependency trees to capture syntactic features. Ruder et al. \cite{Ruder2016} experimented with a hierarchical, bidirectional LSTM model to leverage both intra and inter-sentence relations. Li et al.~\cite{Li2017} proposed an LSTM-based multi-task learning framework.  He et al. \cite{He} used an attention mechanism to focus more on aspect-related words while de-emphasizing aspect-irrelevant words. There exists also an interesting model employing two types of pre-trained embeddings: general-purpose embeddings and domain-specific embeddings~\cite{double-embeddings}.

\subsection{Text Embedding methods for Deep Learning}

Nowadays, many deep learning models in NLP use word embeddings as input features \cite{Ruder2016, Poria2016, double-embeddings}. One of the first word embedding methods is called  Word2Vec \cite{Le2014}. This neural network-based model predicts the target word from its context words (''phone has the best \_ of all available phones'', where \_ denotes the target word ''screen'') or the context words given the target word. A second widely used word embedding is Global Vector (GloVe) \cite{Pennington14glove:global}, which is trained based on a~global word-word co-occurrence matrix. A third technique is fastText \cite{bojanowski2017enriching}. It is based on the Skip-gram model, where each word is represented as a~bag of character n-grams. Researchers started to train and also use domain-dependent word embeddings, e.g., based on product reviews \cite{Poria2016, double-embeddings}. An interesting approach was proposed by Yin et al.~\cite{Yin:2016:UWD:3060832.3061038}, they used dependency paths to generate embeddings.

Lately, the natural language community has focused on more contextual representation. Peters et al. \cite{elmo} proposed deep contextualized word representations (they called the model ELMo). This word embedding technique creates vector space using bidirectional LSTMs trained on a language modeling objective. Then whole language models have appeared and developed rapidly. One of the most well-known is BERT (namely Bidirectional Encoder Representations from Transformers) that uses a modified objective for language modeling called ''masked language modeling''. This model randomly (with some small probability) replaces some words in a sentence with a mask token. Then, a~transformer-based architecture is used to generate a prediction for the masked word based on the unmasked words surrounding it, both to the left and right. Radford et at. \cite{gpt2} moved the normalization layer to the input of each sub-block, and they added a normalization layer after the final self-attention model. Finally, they used a better dataset that emphasizes the diversity of content. Ultimately, we want to mention a pre-trained language model called XLNet that aimed to improved BERT by introducing a variant of language modeling called ``permutation language modeling`` \cite{xlnet}. Instead of predicting masked words independently and in a left to right manner as in BERT, the XLNet model predicts target words based on different orders of source words (no strict left to right order). 

Some NLP models besides word embeddings use char-based embeddings (ELMo~\cite{elmo}, Flair~\cite{flair}), byte-level embeddings~\cite{Kenter2018}, or ngram embeddings\cite{bojanowski2017enriching}. This kind of embedding has been found useful for morphologically rich languages and to deal with the out-of-vocabulary (OOV) problem for tasks, including, in part-of-speech (POS) tagging\cite{DosSantos:2014:LCR:3044805.3045095}, language modeling \cite{Ling2015}, dependency parsing \cite{Ballesteros2015} or named entity recognition \cite{Lample2016}. Zhang et al. \cite{Zhang2015} presented one of the first approaches to sentiment analysis with char embedding using convolution networks. 

\subsection*{}

To the best of our knowledge, this is one of the first paper that reports the use of LSTM-based character embeddings as the extension of the word embedding layer for Long Short Term Memory networks in the Aspect Term Extraction task. It is also the most comprehensive comparison of many combinations of word embeddings, character embeddings, and various variants of LSTMs.

\section{Aspect Term Extraction in Sentiment Analysis}
\label{sec:method_descripti}

This section provides a brief yet formal description of underlying tasks and concepts that concern aspect extraction in sentiment analysis. 

\subsection{Aspect Term Extraction Problem}

% The goal of Aspect Term Extractor is to identify aspects expressed in the documents. Figure \ref{fig:sequence-tagging-example} presents document example with highlighted aspect terms \textit{battery} and \textit{screen}. We want to extract aspects for all processed documents.

% \begin{figure}[!ht]
% \centering
% \includegraphics[scale=0.7]{imgs/sequence-tagging-example.png}
% \caption{Sequence Tagging example with highlighted aspects. We didn't highlight NON-ASPECT words for readability purposes.}
% \label{fig:sequence-tagging-example}
% \end{figure}

Formally, given a collection of documents $D = {d_{1}, ..., d_{n}}$ we want to extract aspects $\mathcal{A} = {A_{1}, ..., A_{n}}$ for each of $n$~documents.

\begin{definition}[Aspect Term Extraction]
Aspect Term Extraction extracts generates pairs $<d_{i}, A_{i}> \in D x \mathcal{A}$ for each of document in the corpus, where $A_{i}$ is a list of aspects for every document.
\end{definition}

Many Aspect Term Extraction models use a sequence tagging approaches.
In sequence tagging, the output of model is a sequence of tags $t = (t_{1}, ..., t_{m})$, also called sequence of labels, and it corresponds to an example sequence $w = (w_{1}, ..., w_{m})$. In NLP, we often use word $w$ sequences as an example sequences and Part-of-Speech tags, entity types, or aspects in aspect-based sentiment analysis as tags $t$. 

\begin{definition}[Sequence Tagging]
The sequence tagger assigns a tag $t$ to each word $w$ as $<w_{i}, t_{i}> \in W x T$, where $W$ is a domain of words and $T = {t^{1}, ..., t^{|T|}}$ is a predefined set of tags. Finally, we get an ordered sequence of word and tag pairs $(w_{1}, t_{1}), (w_{2}, t_{2}), ..., (w_{m}, t_{m})$, where $m$ is the length of the sequence. 
\end{definition}

The Aspect Term Extractor maps each document $d$ into sequence of words $w$, applies sequence tagging model to tag each word as aspect or not $<w_{m}, t_{m}>$, and finally derives document-aspects pairs $<d_{n}, A_{n}>$.

A careful study of the outputs of state-of-the-art Aspect Term Extraction models provided us with valuable insights. It highlighted the need to analyze more profoundly the performance of standard text embeddings and neural network architectures. We wanted to know how the performance of chosen models changes over different text embeddings. Moreover, we tried to figure out what potential mitigation of out-of-vocabulary problems could be obtained for underrepresented and strongly inflected languages such as Polish. We extended the LSTM-based models with character embeddings to evaluate how and under which conditions character embedding could be a useful enhancement for an Aspect Term Extractor.

\subsection{Aspect Term Extraction models}

\begin{figure}[h!]
\centering
\begin{subfigure}{.5\textwidth}
  \centering
  \includegraphics[scale=0.2]{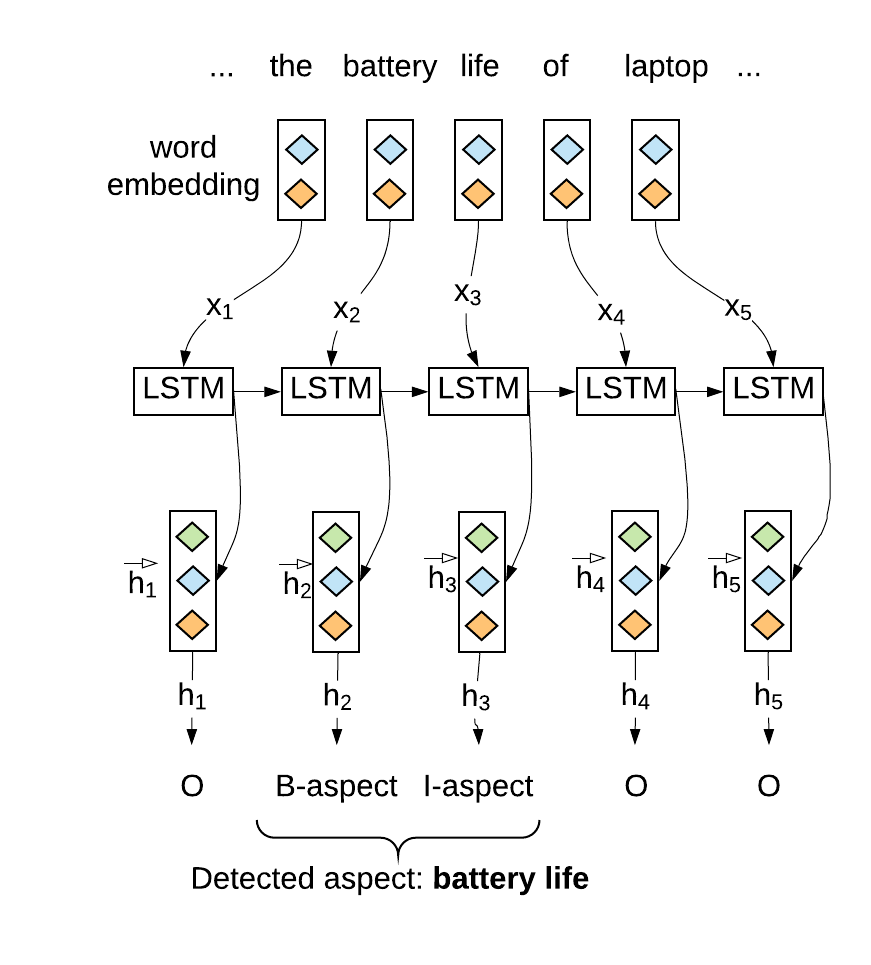}
  \caption{Architecture for word embeddings with simple LSTM.\label{fig:word-lstm}}
\end{subfigure}%
\begin{subfigure}{.5\textwidth}
  \centering
  \includegraphics[scale=0.2]{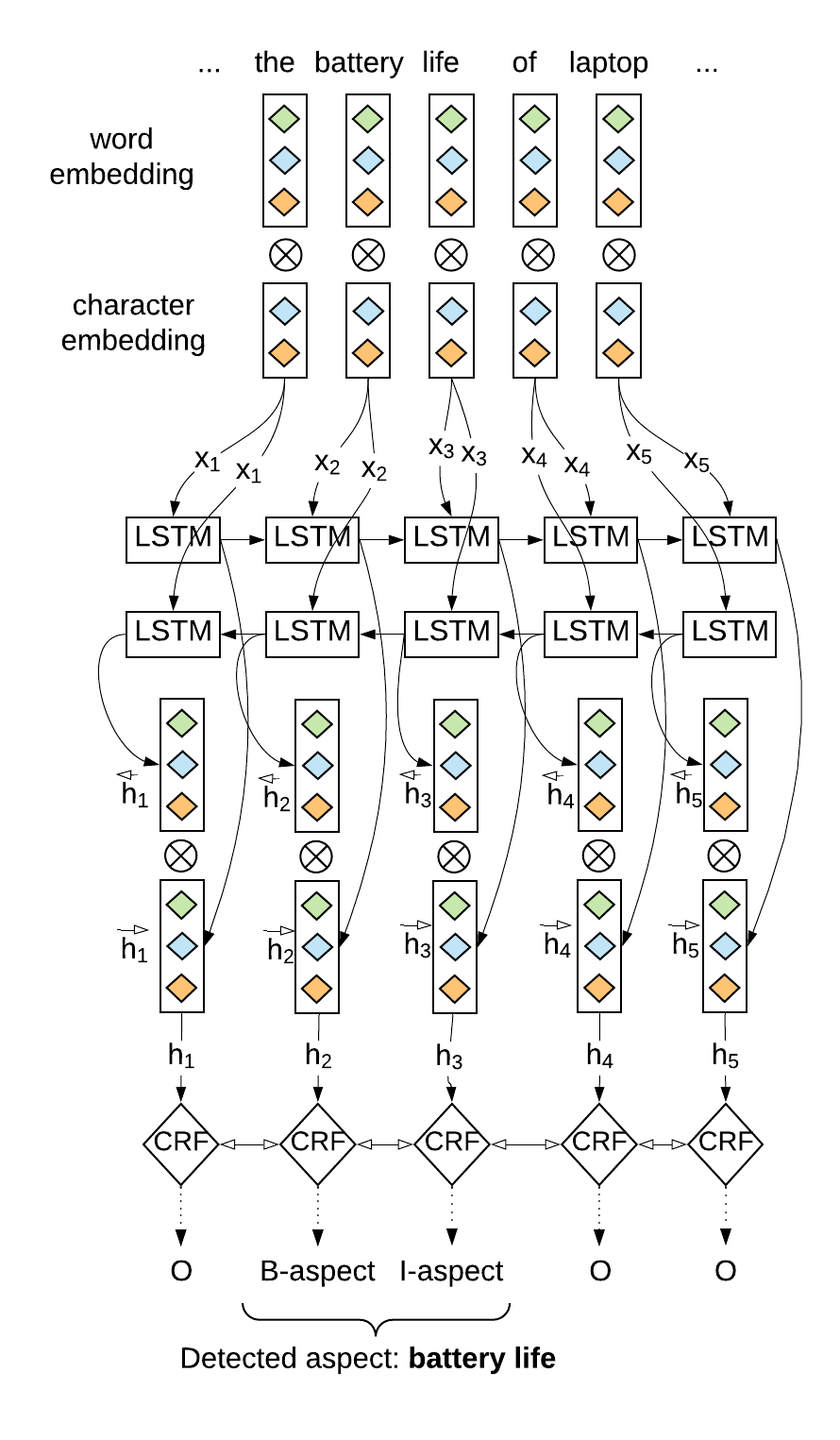}
    \caption{Architecture for word and character embedding with BiLSTM and CRF layer.\label{fig:word-char-bilstm}}
\end{subfigure}
\caption{Example of architectures used in the experiments.}
\label{fig:models}
\end{figure}

In the literature for the Aspect Text Extraction task, two approaches achieve the highest performances. They are based on either Convolution Neural Networks (CNNs) or Long Short-Term Memory Networks (LSTMs). We wanted to focus on generic and straightforward architecture and analyze its performance according to different embedding layers rather than create a very sophisticated customization of neural network architecture. We decided to use LSTM-based models because it is known that Recurrent Neural Networks (RNNs) work very well for sequential data such as in language \cite{Sutskever:2014:SSL:2969033.2969173}. We experimented with standard LSTMs, their bidirectional extension (BiLSTM), and the Conditional Random Field layer on the top of neural network architectures. We experimented with various configurations of the model from the simplest one (Figure \ref{fig:word-lstm}) to the most complex (Figure \ref{fig:word-char-bilstm}). 

Table \ref{table:all_experiments} presents all customizations of our models. As we can see, eight different configurations of features and neural networks were tested. 

\begin{table}[ht!]
\caption{All models used in our experiments. Word and Char denote the word embedding and character embedding, respectively.}
\label{table:all_experiments}
\centering
\begin{tabular}{l|c|c|c}
\hline
\textbf{Model abbreviation} & \textbf{Word} & \textbf{Char} & \textbf{CRF} \\ 
\hline
Wo-LSTM           & yes            & no             & no          \\
Wo-LSTM-CRF       & yes            & no             & yes        \\ 
WoCh-LSTM      & yes            & yes            & no          \\
WoCh-LSTM-CRF  & yes            & yes            & yes        \\ \hline
Wo-BiLSTM         & yes            & no             & no         \\
Wo-BiLSTM-CRF     & yes            & no             & yes       \\ 
WoCh-BiLSTM    & yes            & yes            & no         \\
WoCh-BiLSTM-CRF& yes            & yes            & yes       \\ 
\hline
\end{tabular}
\end{table}

\subsection{Pre-trained Word Embeddings}
\label{sec:word_embedding}

We used several pre-trained word embeddings as we used pre-trained models in transfer learning. Such an approach enables us to mitigate the problem of training models based on limited training data. Our intuition is that aspect indication words (e.g., most of the time aspects are nouns and noun phrases) should appear in regular contexts in large corpora. Moreover, we wanted to evaluate how the performance of various models changes across different word embeddings with a different number of unique words, embedding vector lengths, and varying ratios of out-of-vocabulary words. Pre-trained word embeddings are the primary source of input for neural architecture used in our experiments.

\newpage

We tested several well-established word embeddings:
\begin{enumerate}
    \item \textbf{word2vec} - protoplast model of any neural word embedding trained on Google News. \item \textbf{glove.840B} - Global Vectors for Word Representation proposed by Stanford NLP Group, trained based on Common Crawl with 840B words.
    \item \textbf{glove.42B} - Global Vectors for Word Representation proposed by Stanford NLP Group, trained based on Common Crawl with 42B words.
    \item \textbf{glove.6B*} - Global Vectors for Word Representation proposed by Stanford NLP Group, trained based on Wikipedia 2014 and Gigaword\footnote{\url{https://catalog.ldc.upenn.edu/LDC2012T21}} with 6B words. We used four different word vector lengths: 50, 100, 200, and 300.
    \item \textbf{numberbatch} - Numberbatch consists of state-of-the-art semantic vectors derived from ConceptNet with additions from Glove, Mikolov's word2vec and parallel text from Open Subtitles 2016 \footnote{\url{http://opus.lingfil.uu.se/OpenSubtitles2016.php}} trained via fastText.
    \item \textbf{fastText-wiki-news} - 1 million word vectors (300 dimensions) trained on Wikipedia 2017, UMBC webbase corpus\footnote{\url{https://ebiquity.umbc.edu/resource/html/id/351}} and statmt.org news dataset (16B tokens);
    \item \textbf{fastText-crawl} - 2 million word vectors (300 dimensions) trained on Common Crawl (600B tokens).
    \item \textbf{Amazon Reviews} - word2vec model trained on Amazon Reviews \cite{McAuley:2013:HFH:2507157.2507163}. Since it contains opinionated documents, it should have an advantage over common language texts such as Google News or Common Crawl.
\end{enumerate}

Table \ref{table:pretrained-word-embeddings} shows a summary of all pre-trained word embeddings used in our experiments. As we can see, there are word embeddings that cover more than 2 million unique tokens as well as small vocabulary word embeddings such as sentic2vec (word2vec trained on Amazon Reviews) that contain only 42,000 individual word vectors. We have word embedding representatives trained on general texts such as Wikipedia or Common Crawl and more domain specific text sources as ConceptNet graph or Amazon Reviews. We chose these embeddings to provide a comprehensive comparison of various word embedding techniques, different sources of text used to train them, different length of word vectors, and various vocabulary sizes of word embeddings. We used all pre-trained word embeddings for every model described in Table \ref{table:all_experiments}.

\begin{table}[!h]
\caption{All pre-trained word embeddings used in the experiments.}
\label{table:pretrained-word-embeddings}
\centering
\begin{tabular}{l|c|c|c|c}
\hline
Word Embedding      & Source/Main Source                 & \# of words  & Vocab           & reference \\
\hline
Glove.6B*            & Wikipedia 2014              & 6B           & 400K            & \cite{Pennington14glove:global}           \\
Glove.42B           & Common Crawl                & 42B          & 1.9M            & \cite{Pennington14glove:global}           \\
Glove.840B          & Common Crawl                & 840B         & 2.2M            & \cite{Pennington14glove:global}           \\
word2vec            & Google News                 & 100B         & 3M              & \cite{Le2014}           \\
numberbatch         & ConceptNet 5                & 2M           & 500K            & \cite{speer-conceptnet}          \\
fastText-crawl            & Common Crawl                & 600B         & 2M              & \cite{mikolov2018advances}            \\
fastText-wiki-news            & Wikipedia 2017, news                & 16B         & 1M              & \cite{mikolov2018advances}            \\
sentic2vec          & Amazon Reviews              & 4.7B         & 42K             & \cite{Poria2016}          \\
\hline
\end{tabular}
\caption*{* we used 4 different word vector lengths, to be exact 50, 100, 200 and 300.}
\end{table}

\subsection{Character-based Word Embeddings}

An important distinction of our work from most previous approaches to Aspect Term Extraction is the inclusion and analysis of character-based word embeddings. We measured their impact on the performance of all models across all pre-trained word embeddings. We chose LSTM-based encoding for character embeddings rather than convolutional neural networks because CNNs discover mostly position-invariant features~\cite{Lample2016}. This is usable for image recognition - an object can be spotted anywhere in a~picture, but for NLP tasks order of characters or words are very important, e.g., prefixes and suffixes can convey essential distinctions. Languages with rich inflectional morphology exhibit lexical data sparsity \cite{morphological-smoothing}. The word used to express a~given concept will vary with the syntactic context. Hence, it is unlikely to spot all inflections of a~given lemma, even using large corpora to train word embedding. The character embedding model could mitigate such problems. In our experiments, we focused on static pre-trained word embedding and we tried to add a simple as possible model to mitigate out-of-vocabulary problems that could be learned during the training time of the whole classifier. Hence, we didn't involve any large character-based model such as Flair~\cite{flair} (trained based on a 1-billion word corpus~\cite{chelba2013billion}).

\subsection{Aspect coding in sentence tagging task}
\label{sec:iob}

In our experiments, we used the IOB format for sequence tagging, a.k.a BIO \cite{iob}. It is a~widely used coding scheme for representing sequences. IOB is short for \textit{inside, outside, beginning}. The \textit{B-} prefix before a~tag (i.e., \textit{B-aspect}) indicates that the tag (aspect) is the beginning of the annotated chunk. The \textit{I-} prefix before a~tag (i.e., \textit{I-aspect}) indicates that the tag (aspect) is inside the chunk. \textit{I-tag} could be preceded only by \textit{B-tag} or other \textit{I-tag} for ngram chunks. Finally, the \textit{O} tag (without any tag information, no tag) indicates that a~token does not belong to any of the annotated chunks.

An example sentence \textit{"I charge it at night and skip taking the cord with me because of the good battery life".} is encoded with IOB to: \textit{
I::O
charge::O
it::O
at::O
night::O
and::O
skip::O
taking::O
the::O
cord::B-aspect
with::O
me::O
because::O
of::O
the::O
good::O
battery::B-aspect
life::I-aspect
.::O
}.

\section{A proposal of LSTM-based character embedding model for aspect based sentiment analysis}
\label{sec:lstm}

In this section, we describe our proposal of a Long Short-Term Memory Network architectures used for training Aspect Term Extraction models and character embeddings. Finally, we present a description and justification for using the Conditional Random Fields on top of LSTM architecture. 

\subsection{Long Short-Term Memory Networks}

LSTMs take as input a sequence of vectors (these vectors could represent characters or words) $(x_{1}, x_{2},...,x_{n})$ and return another sequence $(h_{1}, h_{2},...,h_{n})$ that represents some information about this sequence at every step in the input. Nevertheless, the vanilla RNNs are not perfect. The main issue with them is the vanishing gradient problem \cite{Pascanu:2013:DTR:3042817.3043083}. When the network becomes deeper and deeper, the gradients calculated in the back propagation steps become smaller and smaller. Finally, the learning rate slows significantly and long-term dependencies of the language are harder to train. Consequently, RNNs memorize worse and worse words that are far away in the sequence and predictions are biased towards their most recent inputs in the sequence \cite{Bengio:1994:LLD:2325857.2328340}. Long Short-term Memory Networks (LSTMs) have been designed to solve exactly this long-term dependencies using a~memory-cell. This neural network architecture uses special gates in neurons to control the proportion of the input to give to the memory cell, and the proportion from the previous state to forget. LSTMs were proposed by Hochreiter and Schmidhuber \cite{hochreiter1997long} and they are widely used in several different NLP problems \cite{Limsopatham2016, Barnes2017, Ma2017}. To further improve LSTMs and accelerate the training of the network, an extension has been proposed - BiLSTM \cite{DBLP:conf/icassp/2013} - bidirectional LSTM. In this architecture, we split the state neurons of a~regular RNN into two parts - forward and backward. The forward pass $\overrightarrow h_{t}$ is responsible for the positive direction of sequence (e.g., direction according to the word order) and the backward part $\overleftarrow h_{t}$ learns the negative direction (the reverse word order). Finally, the BiLSTM architecture outputs concatenation of vectors from each pass $h_{t} = [\overrightarrow h_{t}; \overleftarrow h_{t}]$. 

We used to following implementation of LSTM:

\begin{equation}
\noindent
    i_{t} = \sigma (W_{i}h_{t-1} + U_{i}x_{t} + b_{i})
\end{equation}

\begin{equation}
    f_{t} = \sigma (W_{f}h_{t-1} + U_{f}x_{t} + b_{f})
\end{equation}

\begin{equation}
    \overline{c_{t}} = \tanh (W_{c}h_{t-1}+U_{c}x_{t} + b_{c}) 
\end{equation}

\begin{equation}
    c_{t} = f_{t} \odot c_{t-1} + i_{t} \odot \overline{c_{t}}
\end{equation}

\begin{equation}
    o_{t} = \sigma (W_{o}h_{t-1}+U_{o}x_{t} + b_{o})
\end{equation}

\begin{equation}
    h_{t} = o_{t} \odot \tanh (c_{t})
\end{equation}

where $\sigma$ is the element-wise sigmoid function and $\odot$ is the element-wise product. $x_{t}$ is the input vector (e.g. word or character embedding) at time $t$, and $h_{t}$ is the hidden state vector at time $t$. $U_{i}$, $U_{f}$, $U_{c}$, $U_{o}$ are the weight matrices of different gates for input, $x_{t}$, $W_{i}$, $W_{f}$, $W_{c}$, $W_{o}$ are the weight matrices for hidden state $h_{t}$, and finally $b_{i}$, $b_{f}$, $b_{c}$, $b_{o}$ denote the bias vectors.

\subsection{Character-based Word Embeddings}

\begin{figure}[!ht]
\centering
\includegraphics[scale=0.3]{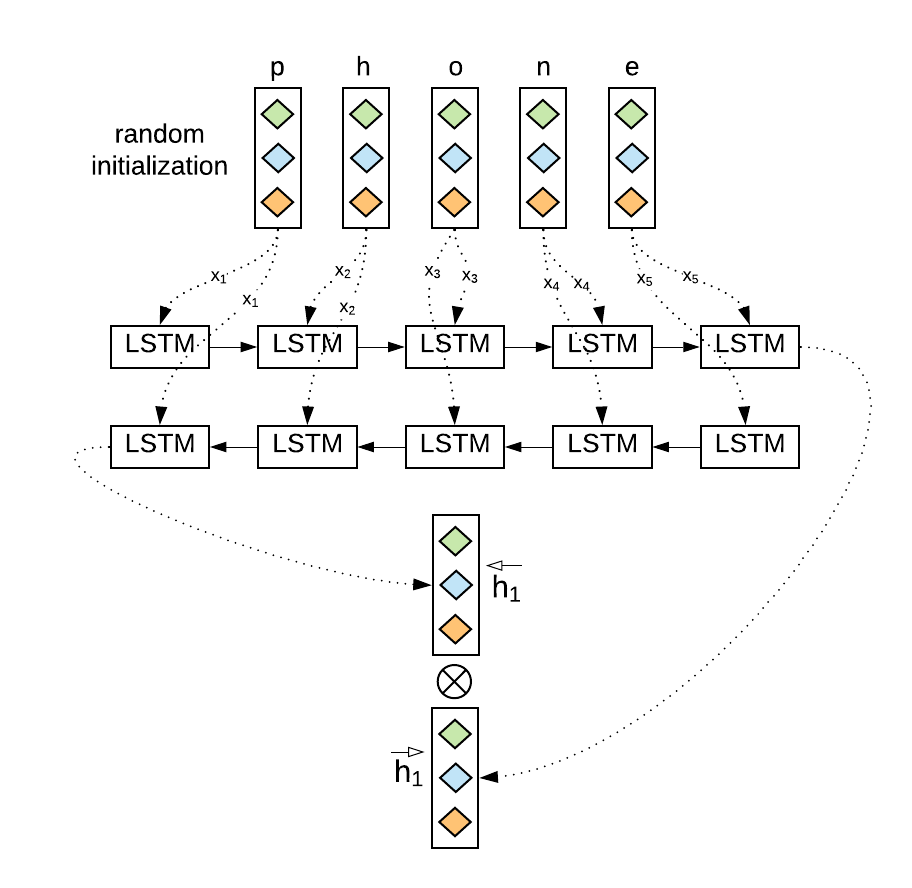}
\caption{Architecture of character embedding.}
\label{fig:char-embedding}
\end{figure}

Figure \ref{fig:char-embedding} presents how we generate embedding for every word using its characters. We take a sequence of character vectors $(x_{1}, x_{2},...,x_{n})$ as an input. The vectors are initialized randomly. Then two passes, forward and backward, generate the output vector representation that is the concatenation of those passes ($\overrightarrow h_{t}$, $\overleftarrow h_{t}$). It is very similar to using BiLSTM for a sequence of words in the sentence. However, in our case we have a sequence of characters in each word. We used such a character-level representation of words in our look-up tables similar to pre-trained word models. 

\subsection{CRF layer}
Conditional Random Field (CRF) is an excellent tool for sequence modeling because it takes into account an object's neighborhood. The LSTM-based models predict tags locally considering only some information about the context. The CRF layer can learn constraints related to the final predicted labels and ensure they are valid.  CRF takes as input a sequence of vectors $z = (z_{1}, z_{2},...,z_{n})$ and returns a sequence of labels $y = (y_{1}, y_{2},...,y_{n})$. $\mathcal{Y}(z)$ is the set of all possible label sequences for $z$. The probabilistic model for CRF defines a family of conditional probability $p(y|z; W, b)$ over possible label sequences of $y$ given $z$ using

\begin{equation}
    p(y|z; W, b) = \frac{\prod_{i=1}^{n} \psi_{i} (y_{i-1}, y_{i}, z)}{\sum_{y' \in \mathcal{Y}(z)} \prod_{i=1}^{n} \psi_{i} (y'_{i-1}, y'_{i}, z)}
\end{equation}

where $\psi_{i}(y', y, z) = \exp (W_{y', y}^{T} z_{i} + b_{y',y})$, $W_{y',y}^{T}$ and $b_{y',y}$ are the weight vector and bias corresponding to label pair $(y', y)$, respectively. 

We used the maximum conditional likelihood estimation for training CRF. For a training set ${\{(z_{i}, y_{i})\}}$, the log-likelihood is given by:

\begin{equation}
    L(W, b) = \sum_{i} \log p(y|z; W, b)
\end{equation}

Maximum likelihood training chooses parameters which maximize the log-likelihood. During the decoding phase we searched for the label sequence $y^{*}$ with the highest conditional probability:

\begin{equation}
    y^{*} = \text{arg}\max\limits_{y \in \mathcal{Y}(z)}\  p(y|z; W,b) 
\end{equation}

\begin{figure}[!ht]
\centering
\includegraphics[scale=0.15]{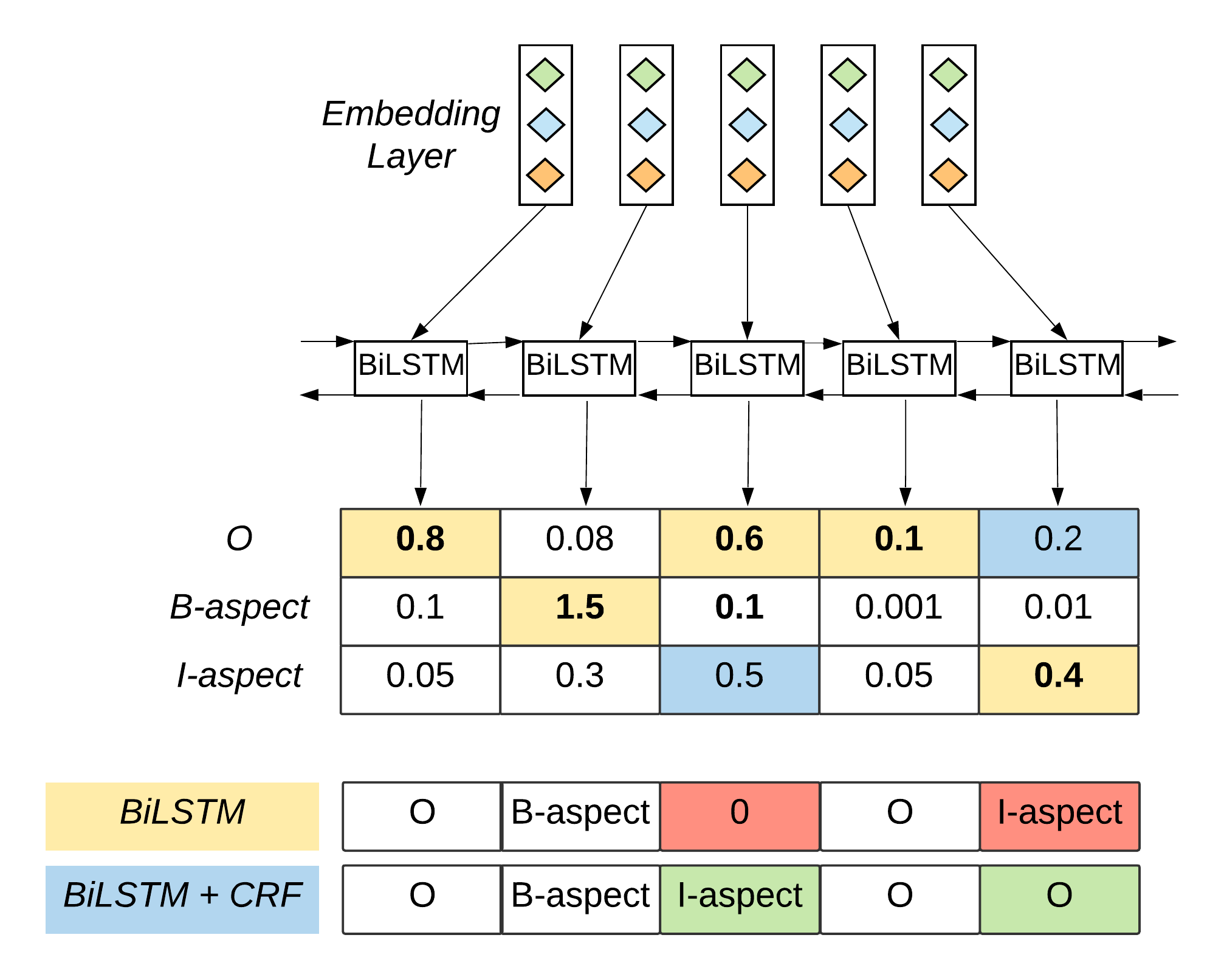}
\caption{BiLSTM outcomes and possible corrections using the CRF layer.\label{fig:bilstm-crf}}
\end{figure}

Figure \ref{fig:bilstm-crf} presents exemplary BiLSTM data extended with the CRF layer. The most interesting part is a~table of BiLSTM predictions that are the input for CRF. We highlighted in light green the highest values (potential predictions). However, some of these predictions are not valid. The CRF layer can use information about previous predictions and choose a~correct tag for words. Thus, we can predict \textit{I-aspect} rather than \textit{O} for the third word and replace the incorrect \textit{I-aspect} tag for the fifth word with no tag. How can we do that? In our case, CRF can learn restrictions or patterns related to the IOB-scheme and tag co-occurrence:

\begin{itemize}
    \item The model predicts  \textit{I-aspect} tags for words that usually look like non-aspect words, such as the second and third word in Figure \ref{fig:bilstm-crf}.
    \item A tag sequence cannot start with \textit{I-aspect} tag. It must begin with either \textit{B-aspect} or no tag - \textit{O}.
\end{itemize}

Hence, we can decrease the number of wrong predictions using the CRF layer at the top of the LSTM-based architecture. In addition, the Conditional Random Field model will be beneficial for multi-word aspects, i.e., \textit{battery life, charging time} in the Laptops domain or names of dishes in the Restaurants domain, i.e., \textit{fish and chips}. 

\section{Experimental setup}
\label{sec:experimental_setup}

We experimented with various sequence tagging approaches for Aspect Term Extraction. All eight considered methods are presented in Table \ref{table:all_experiments}. We used 11 different pre-trained word embeddings. We evaluated a total of 88 combinations of models and text embeddings in the entire experiment. Each combination was run six times to get stable results. In addition, we also ran models with replaced an embedding layer. We used contextual word embedding methods and transformer-based architectures to compare recent state-of-the-art language models as input for LSTM-based architectures.

\subsection{LSTM-based Aspect Term Extraction setup}
\label{sec:lstm-experimental-setup}

We used a grid search to get the best general hyper-parameters based on running four models (namely Wo-LSTM, WoCh-LSTM, Wo-BiLSTM, WoCh-BiLSTM) for all pre-trained word embeddings (\textit{Glove.840B}, \textit{Amazon Reviews}, \textit{fastText-crawl} and word2vec). We chose the most common and the best hyper-parameters from these runs. Finally, we used the following: a mini-batch size equal to 10, maximum sentence length of 30 tokens, word embedding size of 300 (with some exceptions for \textit{Glove.6B} word embedding, see Section \ref{sec:results_word_vectors_len}), and 0.5 as dropout rate \cite{dropout-srivastava}.  We used a single layer for the forward and backward LSTMs whose dimensions are set to 256. Tuning this dimension did not significantly impact model performance. We trained the networks for 25 epochs using cross-entropy. We also tried 50 and 100 epochs, but the results didn't improve after the 25th epoch for any of the experimental scenario. We used the Adam optimizer \cite{adam-optimizer}, and early stopping (max two epochs without improvement). We averaged each model's performance for each pre-trained word embedding across six runs. Our experiments were implemented in keras\footnote{\url{https://keras.io/}} with tensorflow\footnote{\url{https://www.tensorflow.org/}} as backend. The source code for all experiments is available at GitHub\footnote{\url{https://github.com/laugustyniak/aspect_extraction}}.

\subsection{Character-based Word Embedding setup}

We used a single layer for the forward and backward LSTMs whose dimensions are set to 25. Tuning this dimension did not significantly impact model performance. In all our experiments we initialized a random vector of length equal to 25 for each of the characters. The dropout for the input layer was set to 0.5. We concatenated character embeddings of words with word embeddings and fed them together to the network (Figure \ref{fig:word-char-bilstm}). Each of the character embedding models was trained separately on train sets. 

\subsection{SemEval datasets}
\label{sec:datasets}

The SemEval-2014 aspect extraction task consists of customer reviews with annotated aspects of the target entities from two domains: restaurants (3041 sentences) and laptops (3045 sentences). Table \ref{table:semeval} contains statistics of the data provided for each domain. We did not use SemEval 2015 or 2016 aspect extraction datasets because they were prepared as text classification with predefined aspect categories and entities. Moreover, since 2017 there has only been aspect extraction in the tweets challenge. SemEval 2014 consists of sentences with words tagged as aspects. Hence,  SemEval 2014 dataset is the newest, suitable for our sequence tagging approach.

\begin{table}[ht]
\caption{SemEval 2014 dataset profile. Multi-aspect means a~fraction of multi-ngram (two and more words) aspects toward all aspects of the domain.}
\label{table:semeval}
\centering
\begin{tabular}{c|c|c|c}
\hline
\textbf{} & \textbf{} & \textbf{Laptops} & \textbf{Restaurants} \\ 
\hline
\multirow{3}{*}{Train} & \multicolumn{1}{l|}{\# of sentences}       & 3,045    & 3,041 \\
                      & \multicolumn{1}{l|}{\# of aspects}          & 2,358    & 3,693 \\
                      & \multicolumn{1}{l|}{\# of unique aspects}   & 973      & 1,241 \\
                      & \multicolumn{1}{l|}{multi-aspects}          & 37\%     & 25\%  \\ \hline
\multirow{3}{*}{Test}  & \multicolumn{1}{l|}{\# of sentences}       & 800      & 800   \\
                      & \multicolumn{1}{l|}{\# of aspects}          & 654      & 1,134 \\
                      & \multicolumn{1}{l|}{\# of unique aspects}   & 400      & 530 \\
                      & \multicolumn{1}{l|}{multi-aspects}          & 44\%     & 28\%  \\ \hline
\multirow{2}{*}{All}   & \multicolumn{1}{l|}{\# of sentences}       & 3,845    & 3,841 \\
                      & \multicolumn{1}{l|}{\# of aspects}          & 3,012    & 4,827 \\
\hline
\end{tabular}
\end{table}

It is important to highlight some issues related to the annotation process for SemEval 2014 datasets. It was unclear if a noun or noun phrase was used as the aspect term. Aspects referred to the entity as a whole, and not only aspects explicitly mentioned were mismatched \cite{Pontiki2014}. For example, in \textit{this place is awesome}, the word \textit{place} most likely refers to the Restaurant as a whole. Hence, it should not be tagged as an aspect term. In the text \textit{cozy place and good pizza}, it probably refers to the ambiance of the Restaurant that is not explicitly mentioned in the text. In such cases, we would need an additional (external) review context to disambiguate it.

Moreover, there are several reviews rating laptops as such without any particular aspects in mind. This domain often contains implicit aspects expressed by adjectives, e.g., \textit{expensive}, \textit{heavy}, rather than using explicit terms, e.g., \textit{cost}, \textit{weight}. We must remember that in both datasets, annotators were instructed to tag only explicit aspects. 

The majority of the aspects in both datasets are single-words, Table~\ref{table:semeval}. Note that the Laptop dataset consists of proportionally more multi-word aspects than the Restaurant domain. The Restaurant dataset contains many more aspect terms in training and testing subsets, see Table~\ref{table:semeval}. Moreover, it includes more than one aspect per sentence on average. In contrast, the Laptops datasets consist of less than one aspect per sentence on average. 

\begin{table}[!ht]
\caption{The Restaurants and the Laptops datasets: top 20 most frequent aspects.}
\label{table:aspect-examples}
\centering
\begin{tabular}{c|c|c|c}
\hline
Restaurants Train & Restaurants Test & Laptops Train & Laptops Test \\
\hline

\begin{tabular}[c]{@{}l@{}}food\\ service\\ place\\ prices\\ staff\\ menu\\ dinner\\ pizza\\ atmosphere\\ price\\ table\\ meal\\ sushi\\ drinks\\ bar\\ lunch\\ dishes\\ decor\\ ambience\\ portions\end{tabular}   &  \begin{tabular}[c]{@{}l@{}}food\\ service\\ atmosphere\\ staff\\ menu\\ place\\ prices\\ sushi\\ meal\\ drinks\\ waiter\\ price\\ pizza\\ wine\\ waiters\\ desserts\\ lunch\\ dinner\\ chicken\\ bartender\end{tabular}  &  \begin{tabular}[c]{@{}l@{}}screen\\ price\\ use\\ battery life\\ keyboard\\ battery\\ programs\\ features\\ software\\ warranty\\ hard drive\\ windows\\ quality\\ size\\ performance\\ speed\\ applications\\ graphics\\ memory\\ runs\end{tabular} &  \begin{tabular}[c]{@{}l@{}}price\\ performance\\ works\\ os\\ features\\ screen\\ windows 8\\ use\\ size\\ keyboard\\ mac os\\ battery\\ runs\\ battery life\\ speed\\ set up\\ design\\ windows 7\\ usb ports\\ operating system\end{tabular}

\end{tabular}
\end{table}

Analysis of aspect distribution over each dataset appears to be very informative and useful. The top 20 aspect examples according to their frequency from each domain can be found in Table \ref{table:aspect-examples}. On the one hand, aspect terms like \textit{food} and \textit{service} from the Restaurants domain are much more frequent than any other aspect, and, for example, \textit{service} is 4 times more frequent than the third  \textit{place} in the training data. However, aspects in the Laptops domain do not follow this pattern, and they are more balanced.

\subsection{Baseline Methods}

To validate the performance of our proposed models, we compare them against many baselines:

\begin{itemize}
    \item \textbf{DLIREC} \cite{Toh2014}: Top-ranked CRF-based system in ATE sub-task in SemEval 2014 - the Restaurants domain. 
    \item \textbf{IHS R{\&}D} \cite{Chernyshevich2014}: Top-ranked system in ATE sub-task in SemEval 2014 - the Laptops domain.
    \item \textbf{WDEmb} \cite{Yin:2016:UWD:3060832.3061038}: Enhanced CRF with  word  embedding,  linear context embedding and dependency path embedding.
    \item \textbf{RNCRF-O} and \textbf{RNCRF-F} \cite{D16-1059}:  They used tree-structured features and a recursive neural  network  as  the  CRF  input.    \textbf{RNCRF-O} was trained  without  opinion  labels. \textbf{RNCRF-F} was trained with opinion labels  and  some  additional hand-crafted  features.
    \item \textbf{DTBCSNN+F} \cite{10.1007/978-3-319-57529-2_28}: A convolutional stacked neural network using dependency trees to capture syntactic features.
    \item \textbf{MIN} \cite{Li2017}: LSTM-based deep multi-task learning  framework. This  jointly  handles the extraction tasks of aspects and opinions using memory interactions.
    \item \textbf{CNN-Glove.840B} \cite{Poria2016}: deep convolutional neural network using \textit{Glove.840B} word embedding\footnote{This approach was run by us using source code available at~\url{https://github.com/soujanyaporia/aspect-extraction}.}.
    \item \textbf{DE-CNN}~\cite{double-embeddings}: employed a simple CNN model and two types of pre-trained embeddings: general-purpose embeddings and domain-specific embeddings\footnote{This approach was run by us using source code available at~\url{https://github.com/howardhsu/DE-CNN}.}.
\end{itemize}

Besides the baseline models, we also used contextual word embedding methods in our evaluation, namely ELMo \cite{elmo}, BERT \cite{bert}, and Flair \cite{flair}. We tested their performance with the most complex LSTM-based architecture that we used in our experiment, the BiLSTM model, with the CRF layer. Hence, we replaced the embedding layer in our models (static word and character embeddings) with these language models. However, we must remember that most language models are trained on large corpora and using highly demanding architectures (computationally expensive training as well as inference, for example, in the BERT model). We concatenated the token representations from the last four hidden layers of the BERT (bert-large-cased\footnote{\url{https://huggingface.co/transformers/pretrained_models.html}}) model to get the representation for words. The look-up for static word embedding is much faster than the inference of transformer-based language models~\cite{reimers-2019-sentence-bert}. We used the same neural network setup as presented in Section \ref{sec:lstm-experimental-setup}.

\subsection{Quality measure}
\label{sec:quality_measures}

We used several measures to evaluate the quality of the models compared. 

\subsubsection{F1-measure}

The most important measure was the F1-measure (also called F1-score or F-score). This score is the harmonic mean of precision and recall. It ranges between 0 (the worst score) and 1 (the best score). We calculated the F1-measure only for exact matches of the extracted aspects, i.e. the \textit{battery life} aspect will be true positive only when both words have been tagged. It is a strong assumption opposed to some other quality measures with weak F1, when any intersection of words between annotation and prediction are treated as correctly tagged. Hence, the consistency of the annotation process and even one word omitted will impact on the overall performance of the model.  

\subsubsection{Nemeneyi statistical test}

We used the Nemeneyi post-hoc statistical test to find the model groups that differ from each other. Nemeneyi was used on top of the multiple comparison Friedman test \cite{Demsar2006}. The Nemeneyi test makes a pair-wise comparison of all model ranks. We used this test to evaluate models as well as all pre-trained word embeddings. The Nemeneyi test provides a critical distance (CD) for compared groups that are not significantly different from each other as presented in Figure~\ref{fig:statistical-significance-analysis-restaurant-method}. 

\subsubsection{Gain}

We also wanted to evaluate the improvement of some model variations, i.e. the LSTM and BiLSTM architecture. We proposed to calculate the gain - how much method $M_{2}$ gains over method $M_{1}$ - according to Equation~\ref{eq:improvement}: 

\begin{equation}
\label{eq:improvement}
    gain(M_{1}, M_{2}) =\frac{M_{2} - M_{1}}{100\% - M_{1}}
\end{equation}

\noindent where $M_{1}$ and $M_{2}$ denote F1-measures of the first and second methods, respectively.

This equation can be understood as: to what extent does method $M_{2}$ gain within the possible margin left by method $M_{1}$? Interestingly, the one percentage point gained in the F1 measure from 85\% to 86\% is more important ($gain=6.7\%$) than the improvement from 75\% to 76\% ($gain=4\%$), see Figure \ref{fig:char-comparison} for results expressed in $gain$. 

\section{Results and Discussion}
\label{sec:results}

We analyzed 88 method combinations, namely eight model customizations (LSTM vs. BiLSTM, CRF vs. no CRF, word or word with character embeddings) with each of pre-trained word embeddings (Glove, fastText, word2vec, Amazon Reviews, and numberbatch embeddings). We also evaluated the results of three contextual text representations (BERT, Flair, ELMo) using WoCh-BiLSTM-CRF model. We structured the description of the results according to our research questions, making them easier to follow. We added all results in a tabular view in the appendix. However, it would be rather interesting for somebody who wants to investigate the specific case in a very deep manner.

\begin{figure}[ht!]
    \centering
    \begin{subfigure}[b]{0.475\textwidth}
        \centering
        \includegraphics[width=\textwidth]{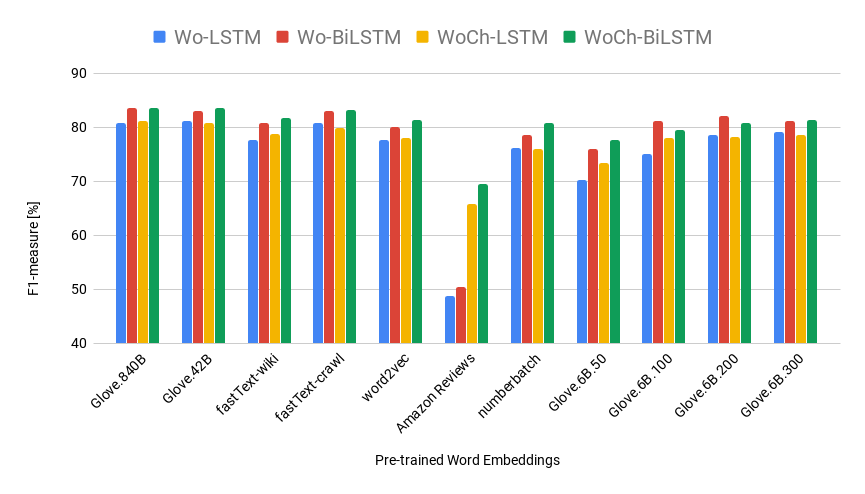}
        \caption[]
        {{\small The Restaurants dataset.}}    
        \label{fig:restaurants-lstm-bilstm}
    \end{subfigure}
    \hfill
    \begin{subfigure}[b]{0.475\textwidth}  
        \centering 
        \includegraphics[width=\textwidth]{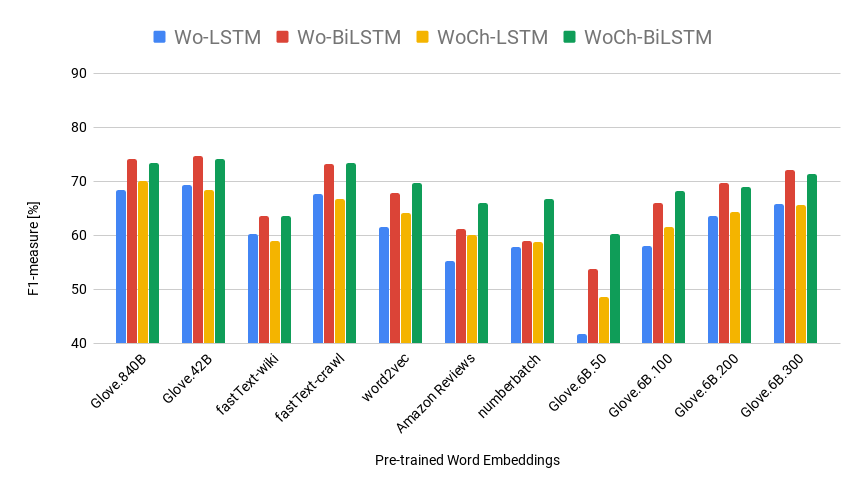}
        \caption[]%
        {{\small The Laptops dataset.}}    
        \label{fig:laptops-lstm-bilstm}
    \end{subfigure}
    
    \caption[]
    {\small Comparison of LSTM and BiLSTM model's performance.}
    \label{fig:lstms-comparison}
\end{figure}

\subsection{Research Question 1 - Robustness}

\textbf{How robust are the general word embeddings in domain-dependent Aspect Term Extraction?} By general embeddings we understand large  pre-trained word representations trained using general texts such as Wikipedia or Common Crawl. 

\subsubsection*{LSTM vs BiLSTM}
\label{sec:results_lstm_vs_bilstm}

There are experiments that confirm the superiority of the BiLSTM-based model over the standard LSTM. This has been verified and demonstrated in Figures~\ref{fig:restaurants-lstm-bilstm} and \ref{fig:laptops-lstm-bilstm}. They contain a~comparison of models with the LSTM and BiLSTM architectures across all evaluated pre-trained word embeddings. Interestingly, we can spot the difference in F1 score distribution for the Restaurant and Laptop datasets. The Restaurant domain scores are flatter and  similar to each other. Most of the time, differences between various word embeddings are not too high. However, the Laptop scores are much more diverse across embeddings. Surprisingly, even well pre-trained models such as \textit{fastText-wiki-news} achieved quite poor performance.

\subsubsection*{Influence of the CRF layer}
\label{sec:results_crf}

\begin{figure}[ht!]
    \centering
    \begin{subfigure}[b]{0.49\textwidth}
        \centering
        \includegraphics[scale=0.22]{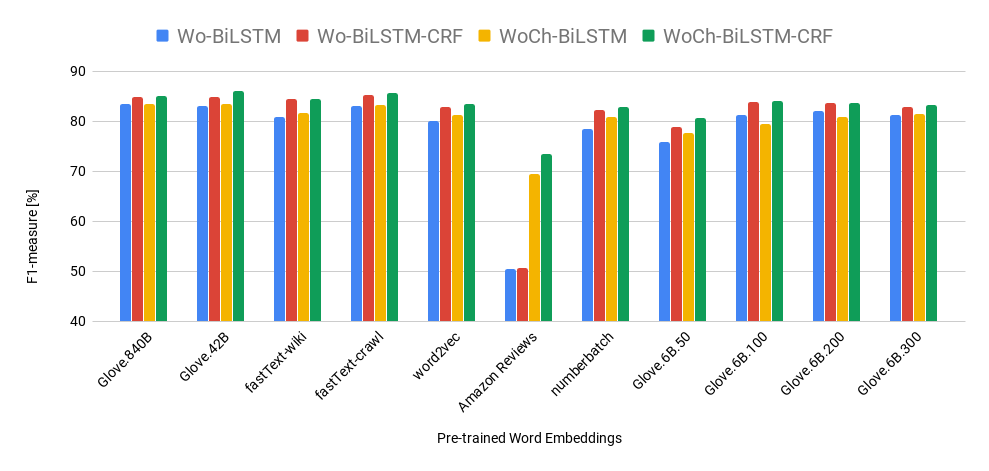}
        \caption[]
        {{\small The Restaurants dataset.}}    
        \label{fig:restaurants-crf}
    \end{subfigure}
    \hfill
    \begin{subfigure}[b]{0.49\textwidth}  
        \centering 
        \includegraphics[scale=0.24]{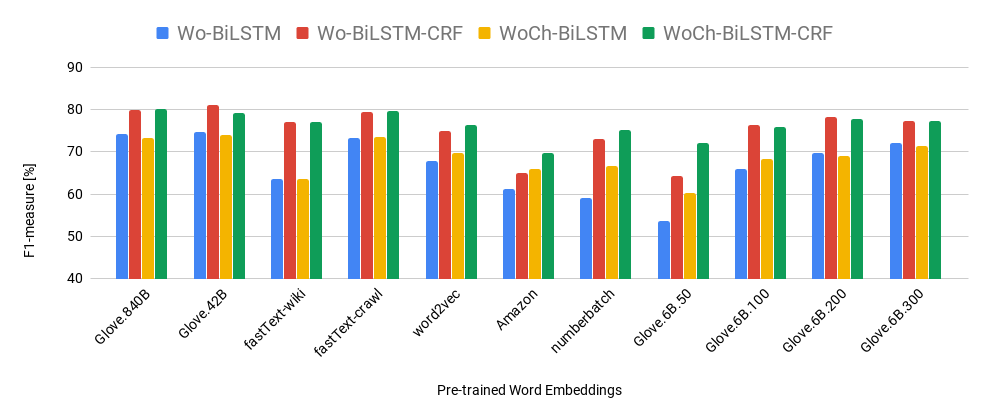}
        \caption[]%
        {{\small The Laptops dataset.}}    
        \label{fig:laptops-crf}
    \end{subfigure}
    
    \caption[]
    {\small The CRF layer extension.}
    \label{fig:crf-comparison}
\end{figure}

The CRF layer added on top of the neural network architecture improves all the models' performance significantly. As we have already mentioned the improvement is higher for the Laptop domain Figure \ref{fig:laptops-crf}. There are some of the word embeddings where using the CRF layer improved the results by more than 10\% percentage points such as for \textit{Glove.6B.50} and surprisingly for \textit{fastText-wiki-news}. The resulting performance of \textit{Glove.6B.50} word embedding was expected. This is a~very short vector representation and it was trained based on a~small corpus. We hypothesized that \textit{fastText-wiki-news} would be reasonably accurate, so we wanted to investigate why such a~lower performance appeared. There are differences between the two fastText models. \textit{fastText-wiki-news} was trained based on Wikipedia and news data, and the \textit{fastText-crawl} was trained using Common Crawl. Moreover, the first model contains one million unique words and the second twice as many. Looking at Figure \ref{fig:words-coverage} we see that better word coverage is presented by \textit{fastText-crawl}. We think that the lower performance of \textit{fastText-wiki-news} would be due to not enough text used to train it. Most of the models with the CRF layer are better than non-CRF approaches. We saw the same pattern for the Laptop dataset (see Figure~\ref{fig:statistical-significance-analysis-restaurant-method}). 

% \begin{figure}[ht!]
% \centering

% \begin{subfigure}[b]{0.85\columnwidth}
% \includegraphics[width=\columnwidth]{imgs/restaurants-crf.png}
% \caption{The Restaurants dataset.}
% \label{fig:restaurants-crf}
% \end{subfigure}

% \begin{subfigure}[b]{0.85\columnwidth}
% \includegraphics[width=\columnwidth]{imgs/laptops-crf.png}
% \caption{The Laptops dataset.}
% \label{fig:laptops-crf}
% \end{subfigure}

% \caption{The CRF layer extension.}
% \label{fig:crf-comparison}
% \end{figure}

\subsection*{Impact of Word Vector Length}
\label{sec:results_word_vectors_len}
\begin{figure}[ht!]
    \centering
    \begin{subfigure}[b]{0.49\textwidth}
        \centering
        \includegraphics[width=\columnwidth]{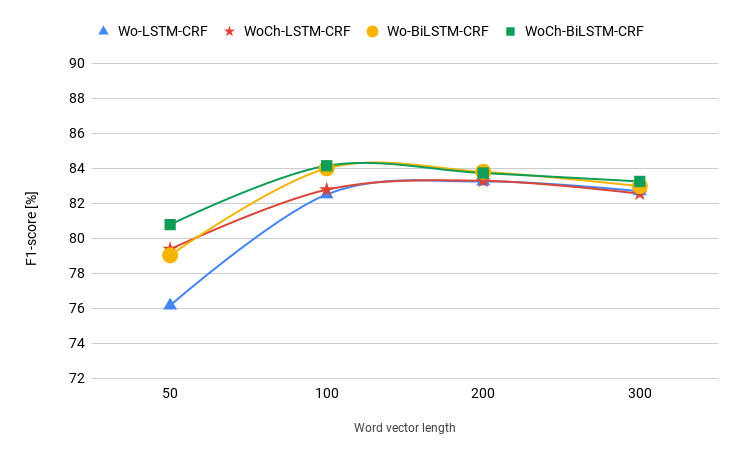}
        \caption[]
        {{\small The Restaurants dataset.}}    
        \label{fig:restaurants-vec-len}
    \end{subfigure}
    \hfill
    \begin{subfigure}[b]{0.49\textwidth}  
        \centering 
        \includegraphics[width=\columnwidth]{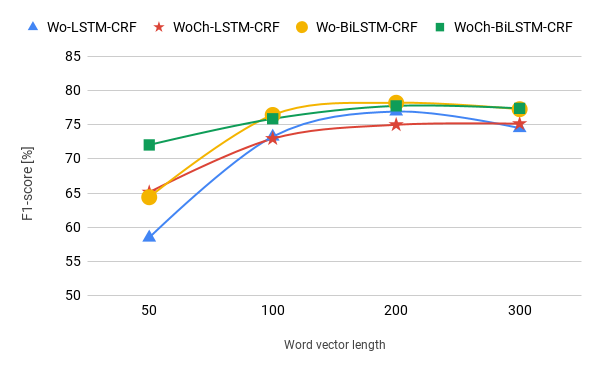}
        \caption[]%
        {{\small The Laptops dataset.}}    
        \label{fig:laptops-vec-len}
    \end{subfigure}
    
    \caption[]
    {\small Comparison of various word vector lengths.}
    \label{fig:vec-len}
\end{figure}

We also evaluated the influence of word vector length on the model's performance. Figure \ref{fig:vec-len} proves that word vector length is important, but the only significant differences can be spotted between length equal to 50 and others. 

% \begin{figure}[ht!]
% \centering

% \begin{subfigure}[b]{0.7\columnwidth}
% \includegraphics[width=\columnwidth]{imgs/restaurants-vector-len.png}
% \caption{The Restaurants dataset.}
% \label{fig:restaurants-vec-len}
% \end{subfigure}

% \begin{subfigure}[b]{0.7\columnwidth}
% \includegraphics[width=\columnwidth]{imgs/laptops-vector-len.png}
% \caption{The Laptops dataset.}
% \label{fig:laptops-vec-len}
% \end{subfigure}

% \caption{Comparison of various word vector lengths.}
% \label{fig:vec-len}
% \end{figure}

Results for \textit{Wo-LSTM-CRF} in the Restaurant domain are equal to 76.2, 82.5, 83.3 and 82.7 for 50, 100, 200 and 300 word vector lengths, respectively. In that case, we can gain more than 6 percentage points using word vector with length equals 100 rather than 50. The improvement for the Laptop dataset was even better than for the Restaurant and achieved almost 15 more percentage points for the \textit{Wo-LSTM-CRF} model. 

It is worth mentioning that when we would like to use word embeddings in production-ready or mobile solutions, it could be a good idea to measure the latency of model inference as well as memory consumption for each worker with a loaded model~\cite{Augustyniak2020}. In such cases, the possibility to use three times smaller models due to memory could be very beneficial: even a drop of model performance of approximately 1 percent point could be worth considering. 

The improvement for longer vector representations is much smaller when the model contains character embeddings. The character extension could not mitigate word vector lengths enough in pre-trained word embeddings and help to represent texts for Aspect Term Extraction better. Hence, it could be a good idea to use straightforward and short vector representations in individual cases but merged with additional knowledge from character embeddings. 

\subsection{Research Question 2 - Coverage}

\textbf{How does the coverage of word embeddings impact the performance of Aspect Term Extraction?} By the coverage, we understand how many unique words from our datasets are not present in vocabularies of the pre-trained word embeddings. This information is strictly related to the out-of-vocabulary problem with any static (based on look-up tables) word representations. We name mentioned coverage also as an out-of-vocabulary ratio (OOV ratio). This research question is one of the most exciting parts of our experiments and a significant starting point if you want to improve quickly your models that are based on static text representations. 

\subsubsection*{Word Embedding Vocabulary Coverage}
\begin{figure}[!h]
\centering
\includegraphics[width=0.6\columnwidth]{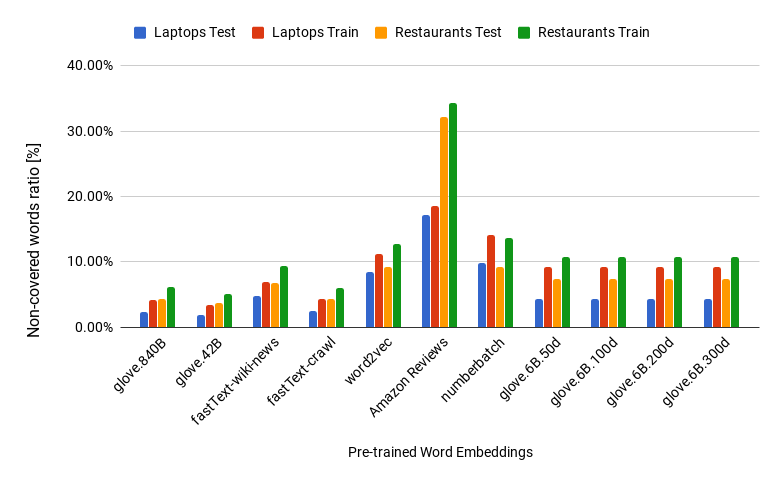}
\caption{Percent of words not covered by each pre-trained word embedding.}
\label{fig:words-coverage}
\end{figure}

In a~deeper analysis of results and influence of different methods, we start with a word coverage comparison between pre-trained word embeddings and datasets. Figure \ref{fig:words-coverage} shows how many words are not covered by each word embedding. 

The best coverage (OOV ratio) would be equal to 0, which means the embedding covers all words in the datasets. As we can see, most of the word embeddings - even though they are derived from the general language corpora - cover the wording of both datasets quite well. \textit{Glove.42B} proves to be the best model lacking, on average only 3.46\% of words across all subsets of the SemEval data. The second and third best models are \textit{Glove.840B} and \textit{fastText-crawl}, which do not cover 4.23\% and 4.24\% of vocabulary, respectively. On the other hand, the lowest coverage of words is provided by \textit{Amazon Reviews} (25.58\%), \textit{numberbatch} (11.68\%) and \textit{word2vec} (10.38\%). \textit{Amazon Reviews} shows how important are domain dependencies in NLP tasks. Not even one out of every three words has a vector representation in the Restaurants domain. This directly impacts the poor performance of this embedding. Such relatively paltry coverage can be an expected result because \textit{Amazon Reviews} do not consist of recipes, ingredients, and cousins' names. Unexpectedly, \textit{Amazon Reviews} do not give as good coverage as we thought for the Laptops domain. However, this domain is closely related to the electronics and Laptop categories in Amazon word embedding. 

\subsection{Research Question 3 - Character-based Word Embeddings}

\begin{figure}
\centering
\begin{subfigure}{.5\textwidth}
  \centering
  \includegraphics[scale=0.27]{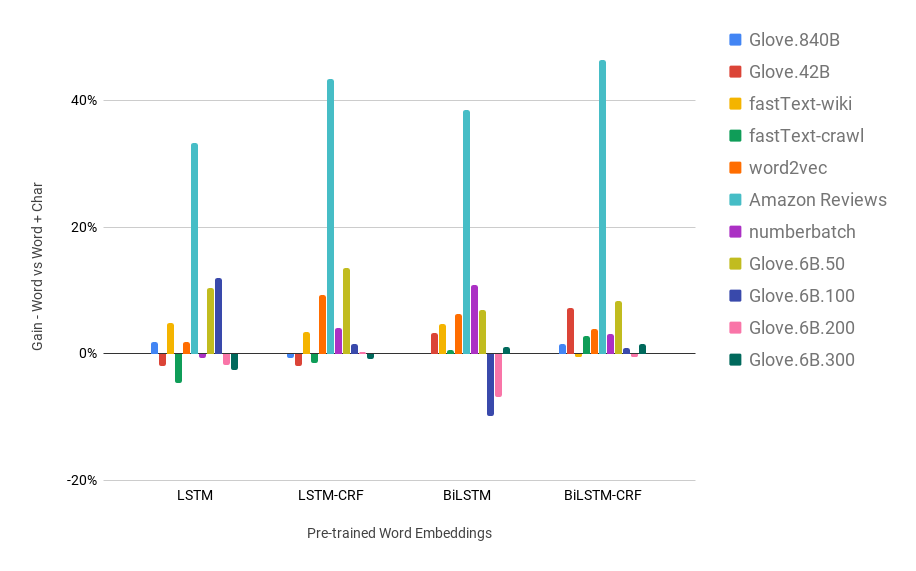}
  \caption{The Restaurant dataset.\label{fig:restaurant-char-extensions}}
\end{subfigure}%
\begin{subfigure}{.5\textwidth}
  \centering
  \includegraphics[scale=0.3]{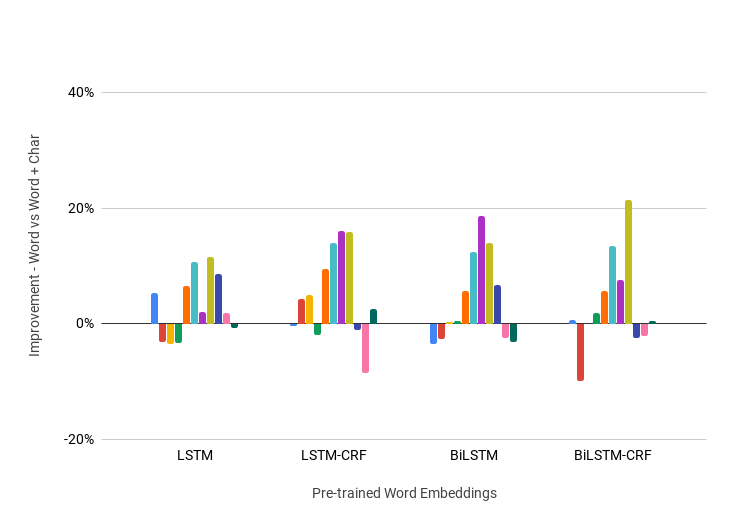}
    \caption{The Laptops dataset.\label{fig:laptops-char-extensions}}
\end{subfigure}
\caption{Gain in F1 measure provided by character extensions of the embedding layer.}
\label{fig:char-comparison}
\end{figure}

\textbf{When character-based word embedding methods are able to eradicate drawbacks of the static pre-trained word embeddings in Aspect Term Extraction?} We calculated and evaluated the influence of extending all neural network architectures with character embeddings according to equation \ref{eq:improvement}. While we analyzed the word coverage between datasets (Table \ref{table:pretrained-word-embeddings}) and pre-trained word embeddings used by us (Figure \ref{fig:char-comparison}) we spotted that the character embeddings work very well for low coverage word embedding such as (\textit{Amazon Reviews} or \textit{ConceptNet numberbatch}). However, character embedding could also add some noise to good word embedding as it is for (\textit{fastText} and \textit{Glove.840B}). Hence, it is essential to understand your dataset and word embedding before applying any character embedding technique. 

\begin{figure}[h!]
    \centering
    \begin{subfigure}[b]{0.475\textwidth}
        \centering
        \includegraphics[width=\textwidth]{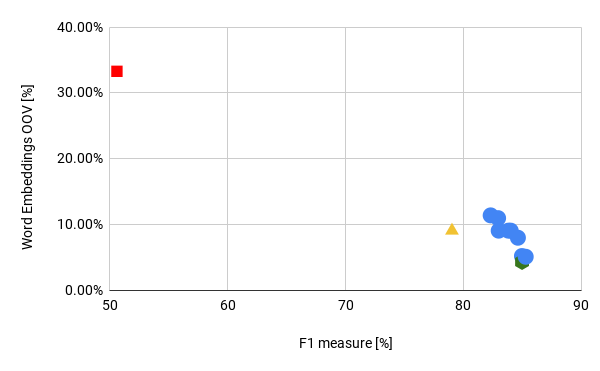}
        \caption[]
        {{\small Restaurants average OOV vs. Wo-BiLSTM-CRF}}    
        \label{fig:restaurants-oov-vs-wo}
    \end{subfigure}
    \hfill
    \begin{subfigure}[b]{0.475\textwidth}  
        \centering 
        \includegraphics[width=\textwidth]{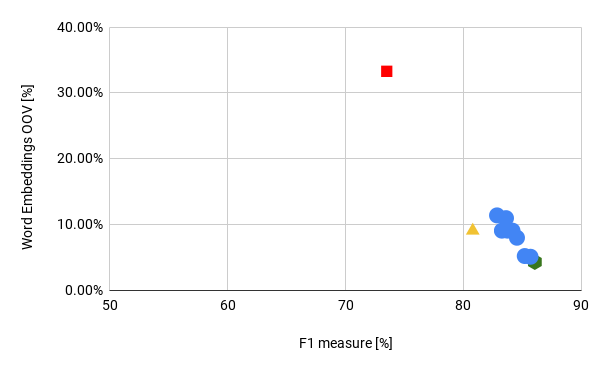}
        \caption[]%
        {{\small Restaurants average OOV vs. WoCh-BiLSTM-CRF}}    
        \label{fig:restaurants-oov-vs-woch}
    \end{subfigure}
    \vskip\baselineskip
    \begin{subfigure}[b]{0.475\textwidth}   
        \centering 
        \includegraphics[width=\textwidth]{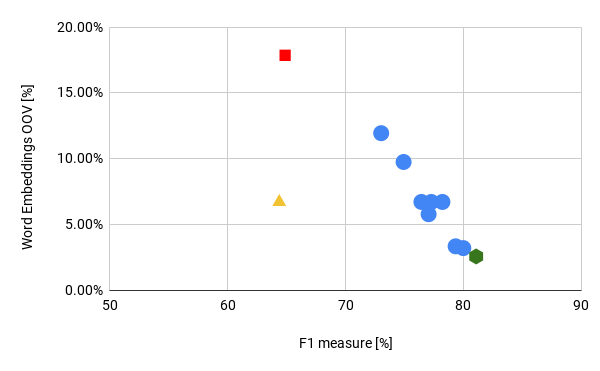}
        \caption[]%
        {{\small Laptops average OOV vs. Wo-BiLSTM-CRF}}    
        \label{fig:laptops-oov-vs-wo}
    \end{subfigure}
    \quad
    \begin{subfigure}[b]{0.475\textwidth}   
        \centering 
        \includegraphics[width=\textwidth]{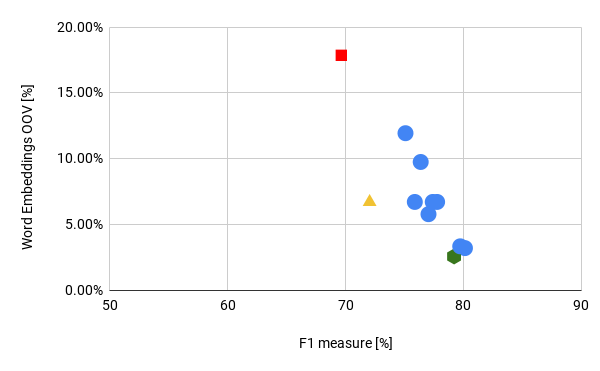}
        \caption[]%
        {{\small Laptops average OOV vs. WoCh-BiLSTM-CRF}}    
        \label{fig:laptops-oov-vs-woch}
    \end{subfigure}
    \caption[]
    {\small Out-of-vocabulary for each of the pre-trained word embeddings and F1 results for our two the best models. Marked pre-trained word embeddings: squares - \textit{Amazon Reviews}, triangles - \textit{Glove.6B}, hexagons - \textit{Glove.42B}} 
    \label{fig:oov-vs-wo}
\end{figure}

Figure \ref{fig:oov-vs-wo} presents the correlation between the out-of-vocabulary ratio and the performance of two models, i.e., Wo-BiLSTM-CRF and WoCh-BiLSTM-CRF. Each point on these scatter plots indicates particular pre-trained word embeddings. There is one particularly interesting example of word embeddings in these graphs. The red squares indicate the OOV ratio and F1 measure of word2vec embeddings trained based on \textit{Amazon Reviews} data. This represents the worst-performing embeddings in all our experiments. The potential improvement by adding character embeddings to these vectors is huge. The second worst word embedding is \textit{Glove.6B}, although its out-of-vocabulary ratio is meager. The problem with this embedding is the length of vectors for each word (length equals only to 50). This case could also be mitigated by adding character embedding to the model as we can see in Figures \ref{fig:laptops-oov-vs-wo} and \ref{fig:laptops-oov-vs-woch}. These figures show an easy to spot correlation. To prove this statement, we calculated the Pearson correlation between the F1-score of various word embeddings for one of the best models  (WoCh-BiLSTM-CRF) for either the Laptop or Restaurant dataset and the corresponding word embedding coverage is very high. The Pearson correlation coefficient equals -0.81 for both datasets.

Finally, there exists one more fascinating outcome. We spotted that, in some cases, adding character embedding to the models can hurt performance. If we check the F1 score for the best pre-trained word embedding in our experiments (\textit{glove.42B}, green hexagons) we see immediately that the performance of this model dropped after adding character embeddings. We hypothesize that the pre-trained representation has already been excellent according to the OOV ratio as well as the size of datasets used to train it. Hence, character-based vectors only added some noise and lowered the model's performance. 

\section{Discussion}

Summarizing, some of the comparisons presented here could be trivial ones. However, in some individual cases, they could provide us with excellent insights on how to create simple, small models that are not much worse than a~couple of times bigger and more complex ones. Nowadays, a~lot of machine learning specialists are using just the most complex and very heavy models that have been proved to achieve state-of-the-art performance. Still, there exist models much simpler with comparable performance. Finally, we want to compare the best models (Wo-BiLSTM-CRF and WoCh-BiLSTM-CRF) used in our experiments with a couple of state-of-the-art architectures and test the performance of BiLSTM-CRF models using contextual and language model-based embeddings. 

\subsection{Overall Results}
\label{sec:overall-results}

We obtained the best F1-measure of 86.05\% for the Restaurant domain using \textit{Glove.42B} pre-trained word embedding extended with character embedding using BiLSTM together with an additional CRF layer. Interestingly, we received the best results of 81.08\% for the Laptop domain without the character embedding extension. Table \ref{table:overall-results} presents a brief comparison of our models and baselines. \textit{Glove.42B} is the best word embedding regarding word coverage in both datasets. The best of our models achieved better performance than the SemEval 2014 winners - \textit{DLIREC} and \textit{IHS R\&D}.

Moreover, the performance of our straightforward models is not significantly worse than the sophisticated state-of-the-art approaches. The two best models in the literature, the CNN with linguistic patterns (CNN-Glove.840B) and the double embeddings approach (DE-CNN) offer better performance than that of our simple model. However, we have not been able to reproduce the author's results on the same machine and the same datasets even with the author's open-source code. We provide the results of our runs in Table \ref{table:overall-results} as well. 
 
\begin{table}[ht!]
\caption{Comparison of F1 scores for SemEval 2014. Boldfaced are the best results in the section. * These approaches were run by us using source code available at \url{https://github.com/soujanyaporia/aspect-extraction} and \url{https://github.com/howardhsu/DE-CNN}.}
\label{table:overall-results}
\centering
\begin{tabular}{l|c|c|c}
\hline
\textbf{Model} & \textbf{Laptops} & \textbf{Restaurants} \\ 
\hline
DLIREC          & 73.78         & \textbf{84.01} \\
IHS R\&D        &\textbf{74.55} & 79.62 \\
\hline
WDEmb           & 76.16         & 84.97 \\
RNCRF-O         & 74.52         & 82.73 \\
RNCRF-F         & 78.42         & 84.93 \\
DTBCSNN+F       & 75.66         & 83.97 \\
MIN             & 77.58         & - \\
CNN-Glove.840B  & \textbf{82.32}         & \textbf{87.17} \\
CNN-Glove.840B* & 77.36*         & 82.76* \\
DE-CNN          & 81.59 & - \\
DE-CNN*         & 78.70*& - \\
\hline
ELMo-BiLSTM-CRF & \textbf{78.81}& \textbf{85.27} \\
BERT-BiLSTM-CRF & 75.74         & 84.10           \\
Flair-BiLSTM-CRF& 77.16         & 85.01\\
\hline
Wo-BiLSTM-CRF-Glove.42B   & \textbf{81.08}   & 84.97 \\ 
WoCh-BiLSTM-CRF-Glove.42B & 79.21            & \textbf{86.05} \\ 
Wo-BiLSTM-CRF-Glove.840B  & 79.99            & 84.96 \\ 
WoCh-BiLSTM-CRF-Glove.840B& 80.13            & 85.2 \\ 
\hline
\end{tabular}
\end{table}

We also compared our simple models against contextual and language model-based embedding approaches using the embedding layer vectors derived from ELMo, BERT, and Flair. All three of these embeddings were used with the BiLSTM-CRF model (we used the same hyper-parameters as for other models, we ran every model six times and got the average F1 score). As we can see, the ELMo embeddings present the best results. However, the F1 score is still below our best models. We probably need to fine-tune these language models for domain data to get better, more competitive representations, but it is not the scope of our paper. We treat these embeddings only as reference values of state-of-the-art embedding methods, and as we can see, they are not magic wands that will always, without any effort, improve your NLP models. They need to be tested and tuned for ones specific case.

We noticed that character embedding could harm the model's performance. For example, the \textit{Wo-BiLSTM-CRF} model with \textit{Glove.42B} word embedding was almost two percentage points better than the same model extended with character embedding. \textit{Glove.840B} yielded a~slightly worse word embedding (in case of word coverage) than \textit{Glove.42B}. Models with \textit{Glove.840B} extended with character embedding prove to be more accurate than the same model alone.

\subsection{Statistical significance analysis}

The Nemeneyi pair-wise test with the Friedman rank test shows the performance across all pre-trained word embeddings and all evaluated methods. As the input for the Nemeneyi test, we used the average value of each model's six runs and embedding combinations. The Nemeneyi analysis provides critical distance (CD, the black horizontal lines on the graphs) for groups of models that are not statistically significantly different from each other.

\begin{figure}[ht!]
\centering

\begin{subfigure}[b]{0.75\columnwidth}
    \includegraphics[width=\columnwidth]{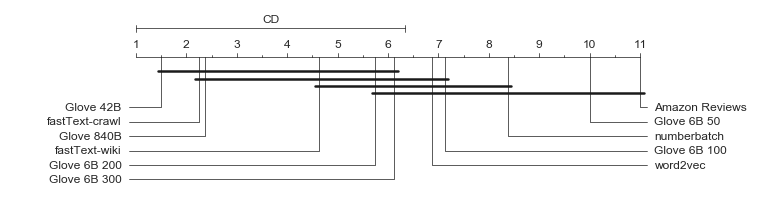}
    \caption{Different pre-trained embeddings across all evaluated methods.}
    \label{fig:statistical-significance-analysis-restaurant-embeddings}
\end{subfigure}

\begin{subfigure}[b]{0.75\columnwidth}
    \includegraphics[width=\columnwidth]{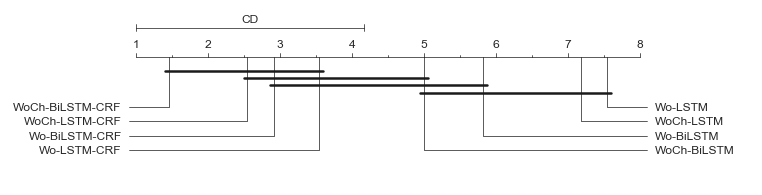}
    \caption{All evaluated methods across pre-trained word embeddings.}
    \label{fig:statistical-significance-analysis-restaurant-method}
\end{subfigure}

\caption{Nemeneyi statistical tests for the Restaurants dataset.}
\label{fig:nemeneyi-restaurants}
\end{figure}

As seen in Figure \ref{fig:nemeneyi-restaurants}  \textit{Glove.42B},  \textit{fastText}, and \textit{Glove.840B} word embeddings are on average the best embedding choice for the Restaurant domain. We can spot a similar pattern for the Laptop domain - Figure \ref{fig:statistical-significance-analysis-restaurant-embeddings}. These three pre-trained word embeddings cover most of the vocabulary contained in the datasets. Interestingly using \textit{Glove.6B} embeddings with word vector length equal to 300 or 200, we can obtain not significantly worse results than for the three best extensive word representation models. This could be important in the productization of machine learning models, where we need to find a~trade-off between accuracy, model loading, and inference time.

\begin{figure}[ht!]
\centering

\begin{subfigure}[b]{0.75\columnwidth}
    \includegraphics[width=\columnwidth]{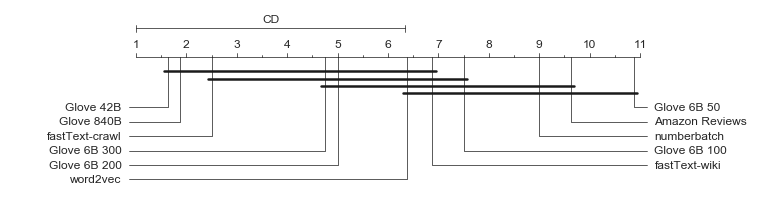}
    \caption{Different pre-trained embeddings across all evaluated methods.}
    \label{fig:statistical-significance-analysis-laptop-embeddings}
\end{subfigure}

\begin{subfigure}[b]{0.75\columnwidth}
    \includegraphics[width=\columnwidth]{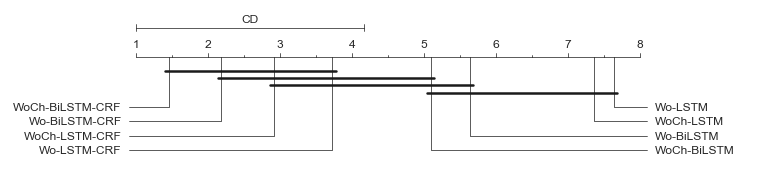}
    \caption{All evaluated methods across pre-trained word embeddings.}
    \label{fig:statistical-significance-analysis-laptop-method}
\end{subfigure}

\caption{Nemeneyi statistical test for the Laptops dataset.}
\label{fig:nemeneyi-laptops}
\end{figure}

The first insight from Figures \ref{fig:statistical-significance-analysis-restaurant-method} and \ref{fig:statistical-significance-analysis-laptop-method} shows a significant improvement for aspect extraction models using CRF as the final layer. All models with a CRF layer prove to have better performance than their equivalents without the CRF layer. The best method \textit{WoCh-BiLSTM-CRF} is always significantly better than any other method without CRF. 

Another valuable outcome of our analysis is related to the performance of \textit{Amazon Reviews} word embedding. Commonly, the domain dependency on natural language processing is critical. However, the efficiency of models based on the \textit{Amazon Reviews} embedding is mediocre at best. This embedding provides a large margin for improvement, Figure \ref{fig:char-comparison}. We hypothesize that this model could be a perfect domain-dependent representation of words for electronics-related data. Maybe we should use this embedding rather in the fine-tuning phase of training, and then it could potentially show more usefulness as in transfer learning approaches~\cite{bartusiak:ENIC2015, BARTUSIAK2019141, Howard2018}.

\section{Conclusions and Future Work}
\label{sec:conclusions_and_future_work}

We performed the first such extensive analysis of sequence tagging approaches for Aspect Term Extraction using various customization of LSTM-based architectures and different word representations. We compared several pre-trained word embeddings and language models, and it can be seen how important the proper embeddings are for the performance of the final model. We must always test which word embedding will work the best for a specific task. For example, \textit{Amazon Reviews} embedding, in theory, should provide us with excellent, domain-dependent representations. However, this performed poorly in our experiments. Besides, even choosing the most recent, very complex language models will not always give us the best performance, and we should select the embedding layer carefully. 

Interestingly under specific conditions, we can use more memory efficient word representations. Word vectors with length equal to 200 or 100 can obtain not significantly worse results than word vectors with a length of 300 or even longer. Plus, we can get a representation that will be a couple of times lighter. This could be valuable in the productization of machine learning models, where we need to find a~trade-off between accuracy, model loading, and inference time. However, most of the time in the literature, researchers and machine learning engineers use standard length representation such as 300 in \textit{Glove.42 or Glove.840B}. 

We also presented evidence that combining word embeddings with character-based representations makes neural architectures more powerful and enables us to achieve better representations, especially for models with a higher OOV ratio. In other words, character-based word representations usually significantly boost embeddings created from the texts with a vocabulary not necessarily well matching the considered domain. However, it must be evaluated for each case separately. We spotted some cases when character embeddings added noise to the models and lowered the overall performance.

As the take-home message, we want to highlight that not always the biggest, the most sophisticated and the most complex model is the best for every case. It is much better and more fun to start experiments with simple models (e.g. one or two layers could be enough) and pre-trained word embeddings (e.g. check the OOV coverage). Then, one should try to improve them quickly and add super easy and fast character embeddings. Character-based word embeddings could help in the task where there may be potentially many OOVs such as concept drifts or new words appearing in social media content. Finally, you can investigate the model's performance, and choose the best method according to your metrics (not always F1). Also, consider other dimensions such as memory, and CPU/GPU consumption. 

Based on our intuition, character embedding should be even more critical for inflected languages such as the Slavic language family. Our future work will focus on the application of the proposed methods to Polish. Moreover, it would also be attractive to fine-tine language models such as BERT based on domain data and then use it to generate better word representations. Another research direction will concentrate on some concepts mentioned above, especially on building particular hierarchies from complex relationships identified between aspects. Finally, we will apply the proposed method for aspect extraction to generate abstractive summaries for various opinion datasets.

\section*{Acknowledgment}
The work was partially supported by the National Science Centre, Poland [grant No. 2016/21/N/ST6/02366,  2016/21/B/ST6/01463] and the European Union's Horizon 2020 research and innovation programme under the Marie Skłodowska-Curie grant agreement [No. 691152] (RENOIR), the Polish Ministry of Science and Higher Education fund for supporting internationally co-financed projects in~2016-2019 (agreement No. 3628/H2020/2016/2) and by the Faculty of Computer Science and Management, Wrocław University of Science and Technology statutory funds.

\newpage
\section*{Appendix}

\newgeometry{margin=7cm} % modify this if you need even more space
\begin{landscape}

\begin{table*}[ht!]
\thispagestyle{empty}
\hskip-4.0cm
\begin{tabular}{c|c|c|c|c|c|c|c|c}
    \hline
\scriptsize{Word Embedding} &                      \scriptsize{Wo-LSTM} &                    \scriptsize{WoCh-LSTM} &                  \scriptsize{Wo-LSTM-CRF} &                \scriptsize{WoCh-LSTM-CRF} &                    \scriptsize{Wo-BiLSTM} &                  \scriptsize{WoCh-BiLSTM} &                \scriptsize{Wo-BiLSTM-CRF} &              \scriptsize{WoCh-BiLSTM-CRF} \\
\hline
Glove.840B     &   80.91 \scriptsize{+/- 1.1} &  81.26 \scriptsize{+/- 0.42} &  85.02 \scriptsize{+/- 0.23} &  84.91 \scriptsize{+/- 0.38} &  83.56 \scriptsize{+/- 0.22} &   83.55 \scriptsize{+/- 0.3} &  84.96 \scriptsize{+/- 0.54} &   85.2 \scriptsize{+/- 0.28} \\
Glove.42B      &   81.28 \scriptsize{+/- 0.6} &  80.91 \scriptsize{+/- 1.33} &  85.64 \scriptsize{+/- 0.28} &  85.37 \scriptsize{+/- 0.57} &  83.08 \scriptsize{+/- 0.37} &  83.64 \scriptsize{+/- 1.04} &  84.97 \scriptsize{+/- 1.22} &  86.05 \scriptsize{+/- 0.37} \\
fastText-wiki-news  &  77.67 \scriptsize{+/- 1.29} &  78.74 \scriptsize{+/- 0.53} &  84.43 \scriptsize{+/- 0.55} &  84.96 \scriptsize{+/- 0.58} &  80.85 \scriptsize{+/- 1.98} &  81.75 \scriptsize{+/- 0.92} &  84.62 \scriptsize{+/- 1.04} &  84.54 \scriptsize{+/- 0.35} \\
fastText-crawl &   80.8 \scriptsize{+/- 1.49} &  79.91 \scriptsize{+/- 1.85} &  85.46 \scriptsize{+/- 0.21} &  85.25 \scriptsize{+/- 0.46} &  83.17 \scriptsize{+/- 0.54} &  83.27 \scriptsize{+/- 0.61} &  85.28 \scriptsize{+/- 0.46} &  85.69 \scriptsize{+/- 0.64} \\
word2vec       &  77.73 \scriptsize{+/- 0.74} &  78.15 \scriptsize{+/- 0.54} &  82.49 \scriptsize{+/- 0.32} &   84.12 \scriptsize{+/- 0.3} &  80.16 \scriptsize{+/- 0.74} &  81.39 \scriptsize{+/- 1.08} &  82.94 \scriptsize{+/- 0.51} &  83.61 \scriptsize{+/- 1.35} \\
Amazon Reviews &  48.78 \scriptsize{+/- 1.02} &   65.81 \scriptsize{+/- 2.3} &  52.09 \scriptsize{+/- 0.98} &  72.84 \scriptsize{+/- 0.62} &  50.49 \scriptsize{+/- 0.87} &  69.53 \scriptsize{+/- 1.52} &   50.63 \scriptsize{+/- 0.5} &   73.5 \scriptsize{+/- 0.91} \\
numberbatch    &  76.26 \scriptsize{+/- 0.75} &   76.11 \scriptsize{+/- 1.9} &  82.19 \scriptsize{+/- 0.84} &  82.92 \scriptsize{+/- 0.33} &  78.57 \scriptsize{+/- 1.04} &  80.89 \scriptsize{+/- 0.26} &  82.31 \scriptsize{+/- 0.47} &  82.85 \scriptsize{+/- 0.41} \\
Glove.6B.50    &   70.24 \scriptsize{+/- 2.9} &  73.35 \scriptsize{+/- 1.63} &  76.16 \scriptsize{+/- 0.77} &   79.38 \scriptsize{+/- 0.4} &   75.97 \scriptsize{+/- 0.7} &  77.64 \scriptsize{+/- 1.33} &  79.03 \scriptsize{+/- 0.64} &  80.79 \scriptsize{+/- 0.42} \\
Glove.6B.100   &  75.04 \scriptsize{+/- 1.81} &  78.04 \scriptsize{+/- 0.69} &  82.52 \scriptsize{+/- 0.46} &  82.79 \scriptsize{+/- 0.51} &   81.3 \scriptsize{+/- 0.28} &  79.47 \scriptsize{+/- 2.27} &  84.01 \scriptsize{+/- 0.49} &  84.16 \scriptsize{+/- 0.34} \\
Glove.6B.200   &   78.69 \scriptsize{+/- 1.3} &   78.3 \scriptsize{+/- 0.81} &  83.26 \scriptsize{+/- 0.33} &    83.3 \scriptsize{+/- 0.2} &   82.09 \scriptsize{+/- 0.7} &  80.87 \scriptsize{+/- 0.85} &  83.81 \scriptsize{+/- 0.19} &  83.74 \scriptsize{+/- 0.56} \\
Glove.6B.300   &  79.22 \scriptsize{+/- 0.58} &   78.7 \scriptsize{+/- 0.54} &   82.7 \scriptsize{+/- 0.78} &  82.56 \scriptsize{+/- 0.66} &  81.31 \scriptsize{+/- 0.61} &   81.5 \scriptsize{+/- 0.68} &  82.99 \scriptsize{+/- 0.53} &  83.26 \scriptsize{+/- 0.53} \\
\hline
\end{tabular}
\caption{All results averaged over 6 runs with std - the Restaurant dataset.}
\label{tab:results-restaurants}
\end{table*}

\begin{table*}[ht!]
\thispagestyle{empty}
\hskip-4.0cm
\begin{tabular}{c|c|c|c|c|c|c|c|c}
\hline
\scriptsize{Word Embedding} &                      \scriptsize{Wo-LSTM} &                    \scriptsize{WoCh-LSTM} &                  \scriptsize{Wo-LSTM-CRF} &                \scriptsize{WoCh-LSTM-CRF} &                    \scriptsize{Wo-BiLSTM} &                  \scriptsize{WoCh-BiLSTM} &                \scriptsize{Wo-BiLSTM-CRF} &              \scriptsize{WoCh-BiLSTM-CRF} \\
\hline
Glove.840B     &  68.38 \scriptsize{+/- 3.61} &  70.09 \scriptsize{+/- 0.61} &  77.72 \scriptsize{+/- 1.42} &  77.66 \scriptsize{+/- 0.46} &  74.25 \scriptsize{+/- 0.87} &  73.38 \scriptsize{+/- 2.46} &  79.99 \scriptsize{+/- 0.72} &  80.13 \scriptsize{+/- 0.34} \\
Glove.42B      &  69.44 \scriptsize{+/- 2.13} &  68.47 \scriptsize{+/- 1.73} &  77.39 \scriptsize{+/- 0.63} &  78.36 \scriptsize{+/- 1.17} &  74.78 \scriptsize{+/- 1.46} &  74.11 \scriptsize{+/- 1.19} &  81.08 \scriptsize{+/- 0.69} &  79.21 \scriptsize{+/- 0.46} \\
fastText-wiki-news  &  60.32 \scriptsize{+/- 4.55} &  58.96 \scriptsize{+/- 2.42} &  74.66 \scriptsize{+/- 1.49} &  75.93 \scriptsize{+/- 0.81} &  63.54 \scriptsize{+/- 4.15} &  63.66 \scriptsize{+/- 4.49} &  77.05 \scriptsize{+/- 2.18} &  77.04 \scriptsize{+/- 2.45} \\
fastText-crawl &  67.75 \scriptsize{+/- 4.05} &  66.71 \scriptsize{+/- 4.88} &  77.95 \scriptsize{+/- 1.79} &  77.53 \scriptsize{+/- 0.93} &  73.32 \scriptsize{+/- 1.32} &  73.44 \scriptsize{+/- 2.77} &  79.34 \scriptsize{+/- 1.23} &  79.73 \scriptsize{+/- 1.36} \\
word2vec       &  61.59 \scriptsize{+/- 2.43} &   64.1 \scriptsize{+/- 2.67} &  72.88 \scriptsize{+/- 1.12} &  75.44 \scriptsize{+/- 1.57} &  67.96 \scriptsize{+/- 2.15} &  69.77 \scriptsize{+/- 2.84} &   74.93 \scriptsize{+/- 1.0} &  76.38 \scriptsize{+/- 1.37} \\
Amazon Reviews &  55.18 \scriptsize{+/- 1.77} &  60.01 \scriptsize{+/- 1.18} &  65.15 \scriptsize{+/- 0.73} &   70.04 \scriptsize{+/- 1.3} &  61.22 \scriptsize{+/- 1.14} &  66.06 \scriptsize{+/- 1.11} &  64.89 \scriptsize{+/- 0.75} &  69.65 \scriptsize{+/- 0.97} \\
numberbatch    &  57.88 \scriptsize{+/- 2.48} &  58.77 \scriptsize{+/- 3.86} &   69.19 \scriptsize{+/- 2.5} &  74.15 \scriptsize{+/- 0.39} &  59.02 \scriptsize{+/- 7.19} &  66.69 \scriptsize{+/- 2.07} &  73.03 \scriptsize{+/- 1.02} &  75.09 \scriptsize{+/- 1.75} \\
Glove.6B.50    &  41.77 \scriptsize{+/- 6.04} &   48.5 \scriptsize{+/- 8.45} &  58.48 \scriptsize{+/- 1.18} &  65.12 \scriptsize{+/- 2.48} &  53.71 \scriptsize{+/- 1.18} &  60.19 \scriptsize{+/- 4.27} &  64.39 \scriptsize{+/- 3.51} &  72.05 \scriptsize{+/- 1.39} \\
Glove.6B.100   &   58.0 \scriptsize{+/- 3.98} &  61.64 \scriptsize{+/- 1.92} &  73.26 \scriptsize{+/- 2.07} &  72.97 \scriptsize{+/- 1.55} &  65.94 \scriptsize{+/- 3.21} &  68.26 \scriptsize{+/- 0.89} &  76.44 \scriptsize{+/- 3.29} &  75.88 \scriptsize{+/- 2.22} \\
Glove.6B.200   &  63.69 \scriptsize{+/- 2.15} &  64.39 \scriptsize{+/- 2.07} &  76.94 \scriptsize{+/- 0.96} &   75.0 \scriptsize{+/- 0.99} &  69.71 \scriptsize{+/- 2.83} &  68.98 \scriptsize{+/- 2.34} &  78.22 \scriptsize{+/- 1.67} &  77.77 \scriptsize{+/- 1.47} \\
Glove.6B.300   &  65.82 \scriptsize{+/- 2.33} &   65.59 \scriptsize{+/- 1.7} &  74.51 \scriptsize{+/- 1.98} &  75.15 \scriptsize{+/- 0.54} &   72.2 \scriptsize{+/- 1.39} &  71.32 \scriptsize{+/- 0.82} &  77.28 \scriptsize{+/- 0.87} &   77.4 \scriptsize{+/- 0.24} \\
\hline
\end{tabular}
\caption{All results averaged over 6 runs with std - the Laptops dataset.}
\label{tab:results-laptops}
\end{table*}

\end{landscape}
\restoregeometry

% \section*{References}

\bibliography{paper}

\end{document}